\documentclass[10pt,twocolumn,letterpaper]{article}

\usepackage{cvpr}
\usepackage{times}
\usepackage{epsfig}
\usepackage{graphicx}
\graphicspath{{figs/}}
\usepackage{amsmath}
\usepackage{amssymb}
\usepackage[percent]{overpic}
\usepackage{upgreek}
\usepackage{textcomp}
\usepackage{gensymb}
\usepackage{booktabs}

\def\eqref#1{equation~\ref{#1}}

\def\1{\bm{1}}

\newcommand{\Var}{\mathrm{Var}}

\usepackage[breaklinks=true,bookmarks=false]{hyperref}

\cvprfinalcopy 

\def\cvprPaperID{6327} 

\ifcvprfinal\fi
\begin{document}

\title{A Poisson-Gaussian Denoising Dataset with Real Fluorescence Microscopy Images}

\author{Yide Zhang\thanks{Equal contribution.}, Yinhao Zhu\footnotemark[1], Evan Nichols, 
Qingfei Wang, Siyuan Zhang, Cody Smith, Scott Howard\\
University of Notre Dame\\
Notre Dame, IN 46556, USA\\
{\tt\small \{yzhang34, yzhu10, enichol3, qwang9, szhang8, csmith67, showard\}@nd.edu}
}

\maketitle


\begin{abstract}
Fluorescence microscopy has enabled a dramatic development in modern biology. Due to its inherently weak signal, fluorescence microscopy is not only much noisier than photography, but also presented with Poisson-Gaussian noise where Poisson noise, or shot noise, is the dominating noise source. To get clean fluorescence microscopy images, it is highly desirable to have effective denoising algorithms and datasets that are specifically designed to denoise fluorescence microscopy images. While such algorithms exist, no such datasets are available. In this paper, we fill this gap by constructing a dataset - the Fluorescence Microscopy Denoising (FMD) dataset - that is dedicated to Poisson-Gaussian denoising. The dataset consists of 12,000 real fluorescence microscopy images obtained with commercial confocal, two-photon, and wide-field microscopes and representative biological samples such as cells, zebrafish, and mouse brain tissues. We use image averaging to effectively obtain ground truth images and 60,000 noisy images with different noise levels. We use this dataset to benchmark 10 representative denoising algorithms and find that deep learning methods have the best performance. To our knowledge, this is the first real microscopy image dataset for Poisson-Gaussian denoising purposes and it could be an important tool for high-quality, real-time denoising applications in biomedical research.
\end{abstract}

\section{Introduction}
\label{sec:intro}

\begin{figure}[t]
\begin{center}
\includegraphics[width=1\linewidth]{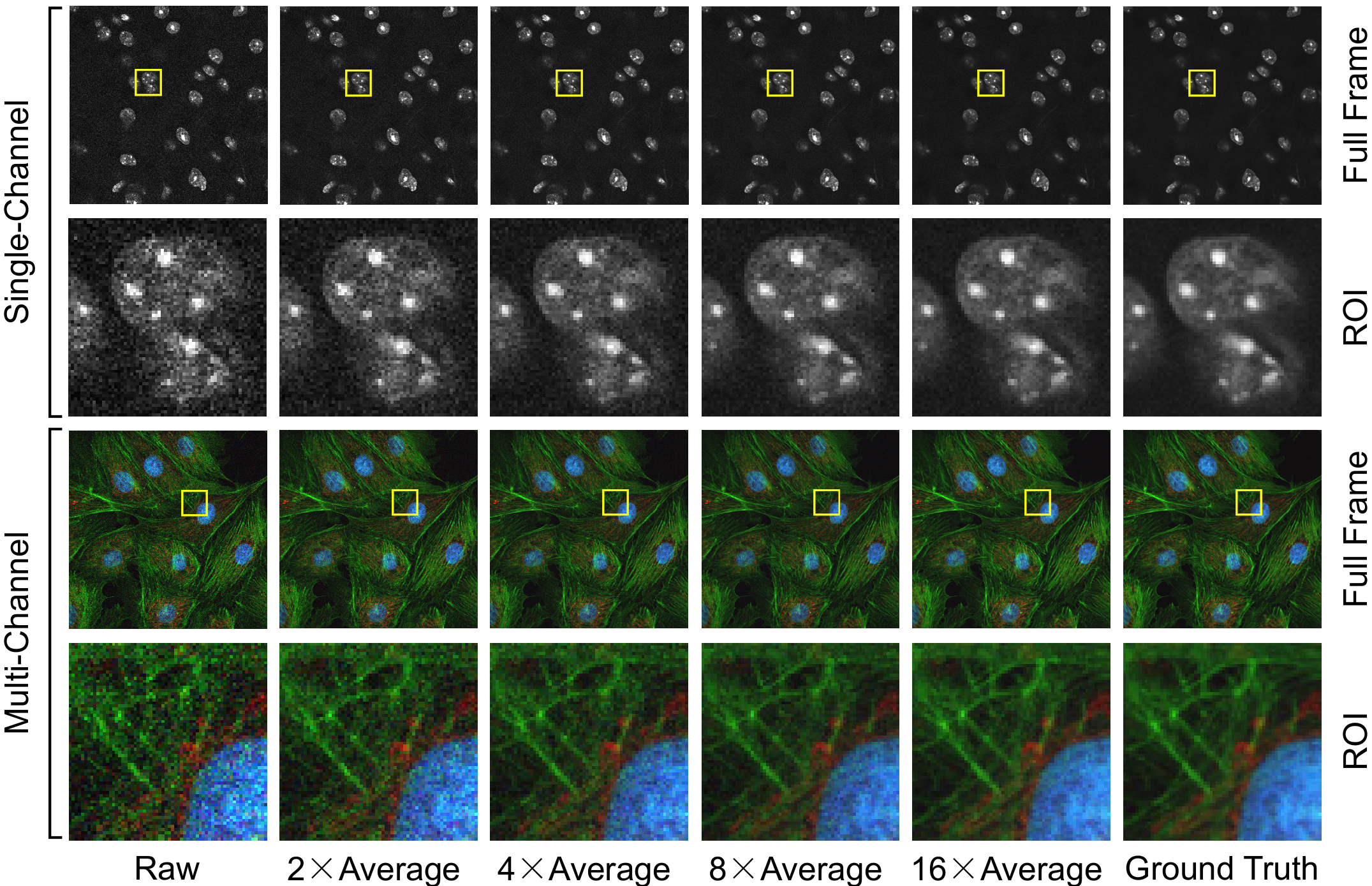}
\end{center}
\caption{Examples of images with different noise levels and ground truth. The single-channel (gray) images are acquired with two-photon microscopy on fixed mouse brain tissues. The multi-channel (color) images are obtained with two-photon microscopy on fixed BPAE cells. The ground truth images are estimated by averaging 50 noisy raw images.}
\label{fig:fig_diff_avg_roi}
\end{figure}

Fluorescence microscopy is a powerful technique that permeates all of biomedical research \cite{lichtman2005fluorescence}. Confocal \cite{pawley2010confocal}, two-photon \cite{denk1990twophoton}, and wide-field \cite{verveer1999widefield} microscopes are the most widely used fluorescence microscopy modalities that are vital to the development of modern biology.
Fluorescence microscopy images, however, are inherently noisy because the number of photons captured by a microscopic detector, such as a photomultiplier tube (PMT) or a charge coupled device (CCD) camera, is extremely weak ($\sim 10^2$ per pixel) compared to that in photography ($\sim 10^5$ per pixel~\cite{morris2015photons}).
Consequently, the measured optical signal in fluorescence microscopy is quantized due to the discrete nature of photons, and fluorescence microscopy images are dominated by Poisson noise, instead of Gaussian noise that denominates in photography~\cite{nam2016holistic}.
One way to obtain clean images is to increase the power of the excitation laser or lamp, but the excitation power is not only limited by the dosage of light a biological sample can receive, but also fundamentally limited by the fluorescence saturation rate; i.e., the fluorescence signal will stop to increase when the excitation power is too high \cite{zhang2017saturation}.
Alternatively, one can get clean images by increasing the imaging time, e.g., pixel dwell time, exposure time, number of line or frame averages; this, however, may cause photodamage to the sample. Moreover, for dynamic or real-time imaging, increasing the imaging time may be impossible since each image has to be captured within tens of milliseconds. Therefore, developing an algorithm to effectively denoise (reduce the noise in) a fluorescence microscopy image is of great importance to biomedical research. Meanwhile, a high-quality denoising dataset is necessary to benchmark and evaluate the effectiveness of the denoising algorithm.

Most of the image denoising algorithms and datasets are created for Gaussian noise dominated images, with a recent focus on denoising with real noisy images, such as smart phones~\cite{abdelhamed2018high} or digital single-lens reflex camera (DSLR) images~\cite{plotz2017benchmarking}. However, there is a lack of a reliable Poisson noise dominated denoising dataset comprising of \textit{real} fluorescence microscopy images. The goal of this work is to fill this gap. More specially, we create a Poisson-Gaussian denoising dataset - the Fluorescence Microscopy Denoising (FMD) dataset - consisting of 12,000 real noisy microscopy images which cover the three most widely used imaging modalities, i.e., confocal, two-photon, and wide-field, as well as three representative biological samples including cells, zebrafish, and mouse brain tissues. With high-quality commercial microscopy, we use image averaging to effectively obtain ground truth images and noisy images with five different noise levels. Some image averaging examples are shown in Figure~\ref{fig:fig_diff_avg_roi}. We further use this dataset to benchmark classic denoising algorithms and recent deep learning models, with or without ground truth. Our FMD dataset is publicly available\footnote{\url{http://tinyurl.com/y6mwqcjs}}, including the code for the benchmark\footnote{\url{https://github.com/bmmi/denoising-fluorescence}}.
To our knowledge, this is the first dataset constructed from real noisy fluorescence microscopy images and designed for Poisson-Gaussian denoising purposes.

\section{Related Work}
\label{sec:related}

There are consistent efforts in constructing denoising dataset with real images to better capture the real-world noise characteristics and evaluate denoising algorithms, such as RENOIR~\cite{anaya2014renoir}, Darmstadt Noise Dataset~\cite{plotz2017benchmarking}, Smartphone Image Denoising Dataset~\cite{abdelhamed2018high}, and PolyU~Dataset~\cite{xu2018real}. Those datasets contain real images taken from either DSLR or smartphones with different ISOs and different number of scenes. The dominating noise in those images is Gaussian or Poisson-Gaussian in real low-light conditions. 
However, there is no dedicated dataset for Poisson noise dominated images, which are inherently different from Gaussin denoising datasets.
This work is dedicated for fluorescence microscopy denoising where the images are corrupted by Poisson-Gaussian noise; in particular, Poisson noise, or shot noise, is the dominant noise source.

Image averaging is the most used method to obtain ground truth images when constructing denoising dataset. The main efforts are spent on image pre-processing, such as image registration to remove the spatial misalignment of an image sequence with the same field of view (FOV)~\cite{alexander2016registration, abdelhamed2018high}, intensity scaling due to the changes of light strength or analog gain~\cite{plotz2017benchmarking}, and methods to cope with clipped pixels due to over exposure or low-light conditions~\cite{anaya2014renoir}. 
The images captured by commercial microscopes in our dataset turns out to be well aligned, and the analog gain is carefully chosen to avoid clipping and to utilize the full dynamic range.

There are two main approaches to denoise an image corrupted by Poisson-Gaussian noise. One way is to directly apply an effective denoising algorithm, such as the PURE-LET method \cite{luisier2011purelet}, which is designed to handle the Poisson-Gaussian denoising problem based on the statistics of the noise model. Another approach is using a nonlinear variance-stabilizing transformation (VST) to convert the Poisson-Gaussian denoising problem into a Gaussian noise removal problem, which is well studied with a considerable amount of effective denoising algorithms to choose from, such as NLM, BM3D, KSVD, EPLL, and WNNM \cite{buades2005non, dabov2007bm3d, aharon2006ksvd, zoran2011learning, gu2014weighted} etc. The VST-based denoising process generally involves three steps. First, the noisy raw images are transformed using a VST designed for the noise model. In our case, we use the generalized Anscombe transformation (GAT) that is designed for Poisson-Gaussian noise \cite{makitalo2013vst}. The VST is able to remove the signal-dependency of the Poisson component, whose noise variance varies with the expected pixel value, and results in a modified image with signal-independent Gaussian noise only and a constant (unitary) noise variance. Next, a Gaussian denoising algorithm is applied to the transformed image. And finally, the Gaussian-denoised data is transformed back via an inverse VST algorithm, such as the exact unbiased inverse transformation \cite{makitalo2013vst}, and the estimation of the noise-free image is obtained.

Recently there is an increasing interest in deep learning based methods for image denoising, where fully convolutional networks (FCNs)~\cite{long2015fully} are used for this image-to-image regression problem. With residual learning and batch normalization, DnCNN~\cite{zhang2017beyond} reports better performance than traditional denoising methods such as BM3D. Further development towards blind image denoising includes incorporating non-uniform noise level map in the input of FFDNet~\cite{zhang2018ffdnet}, or noise estimation network as in CBDNet~\cite{guo2018toward}, or utilizing the non-local self-similarity in UDNet~\cite{lefkimmiatis2017universal} and \cite{plotz2018neural}. These methods all require clean images to supervise the training. There are also progress on denoising methods without paired clean images~\cite{chen2018image} using generative adversarial networks to learn the noise model. In~\cite{lehtinen2018noise2noise}, a Noise2Noise model is trained without clean images at all and outperforms VST+BM3D by almost 2dB on synthetic Poisson noise.

We perform intensive study of the noise statistics of the FMD dataset and show that the noise is indeed Poisson-dominated for two-photon and confocal microscopy, and has larger Gaussian component for wide-field microscopy.
We then benchmark 10 representative denoising algorithms on the FMD dataset, and show better denoising performance with deep learning models than with traditional methods on the real noisy images.

\section{Noise Modeling in Fluorescence Microscopy}
\label{sec:modeling}

The microscopy imaging system is modeled with a Poisson-Gaussian noise model  \cite{foi2008noisemodel, makitalo2013vst}. The model is composed of a Poisson noise component that accounts for the signal-dependent uncertainty, i.e., shot noise, and an additive Gaussian noise component which represents the signal-independent uncertainty such as thermal noise.  
Specifically, let $z_i$, $i=1,2,\cdots,N$, be the measured pixel values obtained with a PMT or a CCD, and
\begin{equation}
    z_i = y_i + n_i = y_i + n_p(y_i)+n_g,
\end{equation}
where $y_i$ is the ground truth and $n_i$ is the noise of the pixel;
the noise $n_i$ is composed of two mutually independent parts, $n_p$ and $n_g$, where $n_p$ is a signal-dependent Poisson noise component that is a function of $y_i$, and $n_g$ is a signal-independent zero-mean Gaussian component.
Denoting $a>0$ as the conversion or scaling coefficient of the detector, i.e., a single detected photon corresponds to a measured pixel value of $a$, and $b\geq 0$ as the variance of the Gaussian noise, we can describe the Poisson and Gaussian (normal) distributions as
\begin{equation}
    \label{eq:dist_ab}
    (y_i+n_p(y_i))/a\sim \mathcal{P}(y_i/a), \quad  n_g\sim \mathcal{N}(0,b).
\end{equation}
Note that $a$ is related to the quantum efficiency of the detector.
Assuming that the Poisson and Gaussian processes are independent, the probability distribution of $z_i$ is the convolution of their individual distributions, i.e.,
\begin{equation}
    p(z_i)=\sum_{k=0}^{+\infty}\left(\frac{\left(\frac{y_i}{a}\right)^k e^{-\frac{y_i}{a}}}{k!}\times\frac{1}{\sqrt{2\pi b}}e^{-\frac{(z_i-a k)^2}{2b}}\right).
\end{equation}
The denoising problem of a microscopy image is then to estimate the underlying ground truth $y_i$ given the noisy measurement of $z_i$.

To denoise a fluorescence microscopy image, one can use algorithms that are specifically designed for Poisson-Gaussian denoising. A more common approach is using VST to stabilize the variance such that the denoising task can be tackled by a well-studied Gaussian denoising method. As a representative VST method, GAT transforms the measured pixel value $z_i$ in the image to
\begin{equation}
    f(z_i)=\frac{2}{a}\sqrt{\max\left(a z_i+\frac{3}{8}a^2+b, 0\right)},
\end{equation}
which stabilizes its noise variance to approximately unity, i.e., $\Var\{f(z_i)\}\approx 1$. A Gaussian denoising algorithm, such as NLM and BM3D, can then be applied to $f(z_i)$ because its noise can be considered as a signal-independent Gaussian process with zero mean and unity variance. Once the denoised version of $f(z_i)$, denoted as $D(z_i)$, is obtained, an inverse VST is used to estimate the signal of interest $y_i$. However, simply applying an algebraic inverse $f^{-1}$ to $D$ will generally result in a biased estimate of $y_i$. An asymptotically unbiased inverse can mitigate the bias, but the denoising accuracy will be problematic for images with low signal levels, a common property of fluorescence microscopy images~\cite{zhang2008poisson}. To address this problem, we use the exact unbiased inverse transformation, which can estimate the signal of interest accurately even at low signal levels~\cite{makitalo2013vst}. In practice, since the exact unbiased inverse requires tabulation of parameters, one can employ a closed-form approximation of it~\cite{makitalo2011close}, i.e.,
\begin{equation}
    \widetilde{I}(D) =\frac{1}{4}D^2+\frac{1}{4}\sqrt{\frac{3}{2}}D^{-1}-\frac{11}{8}D^{-2}+\frac{5}{8}\sqrt{\frac{3}{2}}D^{-3}-\frac{1}{8}.
\end{equation}
The closed-form approximation ensures the denoising accuracy while reducing the computational cost, and the estimated noise-free signal is $\widetilde{y_i}=\widetilde{I}[D(z_i)]$.

To evaluate and benchmark the performances of different denoising algorithms, a ground truth and images with various noise levels are needed, which can be obtained by averaging a series of noisy raw fluorescence microscopy images taken on the same FOV.
In this work, the raw images are the immediate outputs of microscopy detectors, without any preprocessing.
The averaging is performed after ensuring that no image shift larger than a half-pixel can be detected by an image registration algorithm.
Since for different raw images, their Poisson-Gaussian random processes are independent, the average of $S$ noisy raw images, $v^S_i$, can be written as
\begin{align}
    \label{eq:averaging_detail}
    v^S_i&=\frac{1}{S}\sum_{j=1}^{S}z_i^j=\frac{a}{S}\sum_{j=1}^{S}\frac{y_i+n_p^j(y_i)}{a}+\frac{1}{S}\sum_{j=1}^{S}n_g^j\\ \nonumber
    &\sim \frac{a}{S}\mathcal{P}\left(\frac{S y_i}{a}\right)+\frac{1}{S}\mathcal{N}(0,S b),
\end{align}
where $n_p^j$ and $n_g^j$ are the noise realizations of the $j$-th noisy image.
Based on the properties of Poisson and Gaussian distributions, the mean and variance of the averaged image, $v^S_i$, can be written as
\begin{equation}
    \label{eq:averaging_mean_var}
    \mathrm{E}[v^S_i]=y_i, \quad \Var[v^S_i]=\frac{a}{S}y_i+\frac{b}{S}.
\end{equation}
As the number of noisy images used for averaging increases, the noise of ground truth estimation, $\sqrt{\Var[v^S_i]}$, decreases, while the ground truth signal, $\mathrm{E}[v^S_i]$ is invariant; therefore, image averaging is equivalent to increasing the signal-to-noise ratio (SNR) of estimating the ground truth.
We make $S=1,2,4,8,16$ to create images with five different noise levels, and $S=50$ to generate the ground truth. As demonstrated in \cite{alexander2016registration} and also shown in Section~\ref{subsec:estimation_stat}, for fluorescence microscopy images, little image quality improvement can be seen after including around 40 images in averaging.

\section{Dataset}
\label{sec:dataset}

\begin{figure}[t]
\begin{center}
\includegraphics[width=1\linewidth]{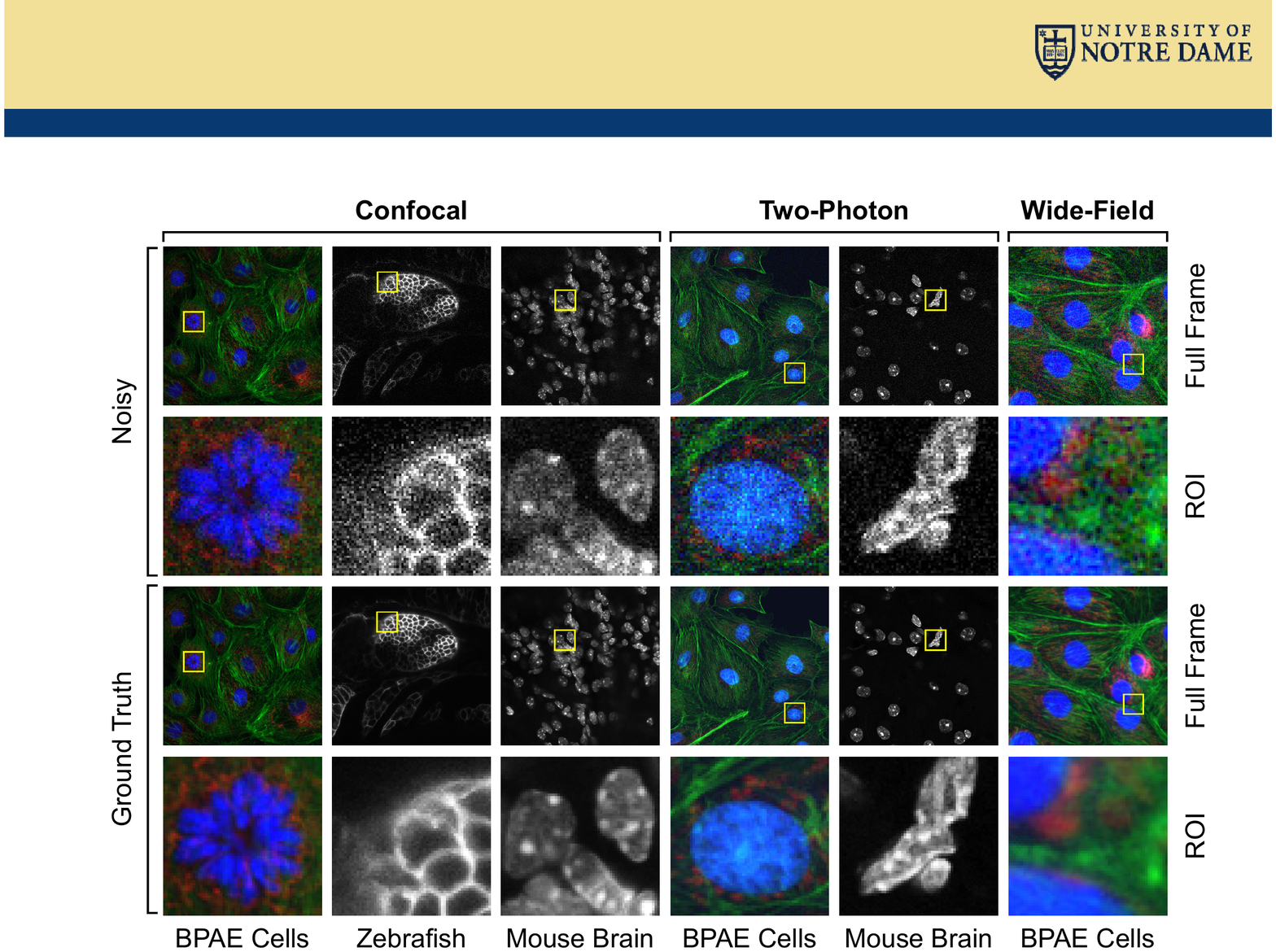}
\end{center}
\caption{Examples of raw fluorescence microscopy images and their estimated ground truth from our FMD dataset. Shown here are FOVs from different microscopy modalities on different biological samples.}
\label{fig:all_mods_with_roi}
\end{figure}

In this Section, we describe the experimental setup that we used to acquire the fluorescence microscopy images. We then discuss how the raw images are utilized to estimate ground truth as well as images with different noise levels. Finally we present the statistics as well as the estimated noise levels of our dataset.

\subsection{Image Acquisition Setup}
\label{sec:image_acquisition}
Our FMD dataset covers the three main modalities of fluorescence microscopy: confocal, two-photon, and wide-field.
All images were acquired with high-quality commercial fluorescence microscopes and imaged with real biological samples, including fixed bovine pulmonary artery endothelial (BPAE) cells [labeled with MitoTracker Red CMXRos (mitochondria), Alexa Fluor 488 phalloidin (F-actin), and DAPI (nuclei); Invitrogen FluoCells F36924], fixed mouse brain tissues (stained with DAPI and cleared), and fixed zebrafish embryos [EGFP labeled \textit{Tg(sox10:megfp)} zebrafish at 2 days post fertilization]. All animal studies were approved by the university's Institutional Animal Care and Use Committee. 

To acquire noisy microscopy images for denoising purposes, we kept an excitation laser/lamp power as low as possible for all imaging modalities. Specifically, the excitation power was low enough to generate a very noisy image, and yet high enough such that the image features were discernible. We also manually set the detector/camera gain to a proper value to avoid clipping and to fully utilize the dynamic range. Although pixel clipping could be inevitable because distinct biological structures with various optical properties could generate extremely bright fluorescence signals that could easily saturate the detector, we were able to maintain a very low number of clipped pixels (less than $0.2\%$ of all pixels) in all imaging configurations. A table summarizing the percentages of clipped pixels to all pixels in the images is presented in the supplementary material.
The details of the fluorescence microscopy setups, including a Nikon A1R-MP laser scanning confocal microscope and a Nikon Eclipse 90i wide-field fluorescence microscope, can also be found in the supplementary material.

For any imaging modality, each sample was imaged with 20 different FOVs, and each FOV was repeatedly captured for 50 times as 50 noise realizations. The acquired images were preprocessed and used for noisy image and ground truth estimation as described in Section~\ref{subsec:noisy_gt_estimate}. Figure~\ref{fig:all_mods_with_roi} shows some example images of a single FOV from different imaging modalities and different samples.

\subsection{Noisy Image and Ground Truth Estimation}
\label{subsec:noisy_gt_estimate}
\paragraph{Image registration} 

\begin{figure}[t]
    \centering
    \includegraphics[width=\linewidth]{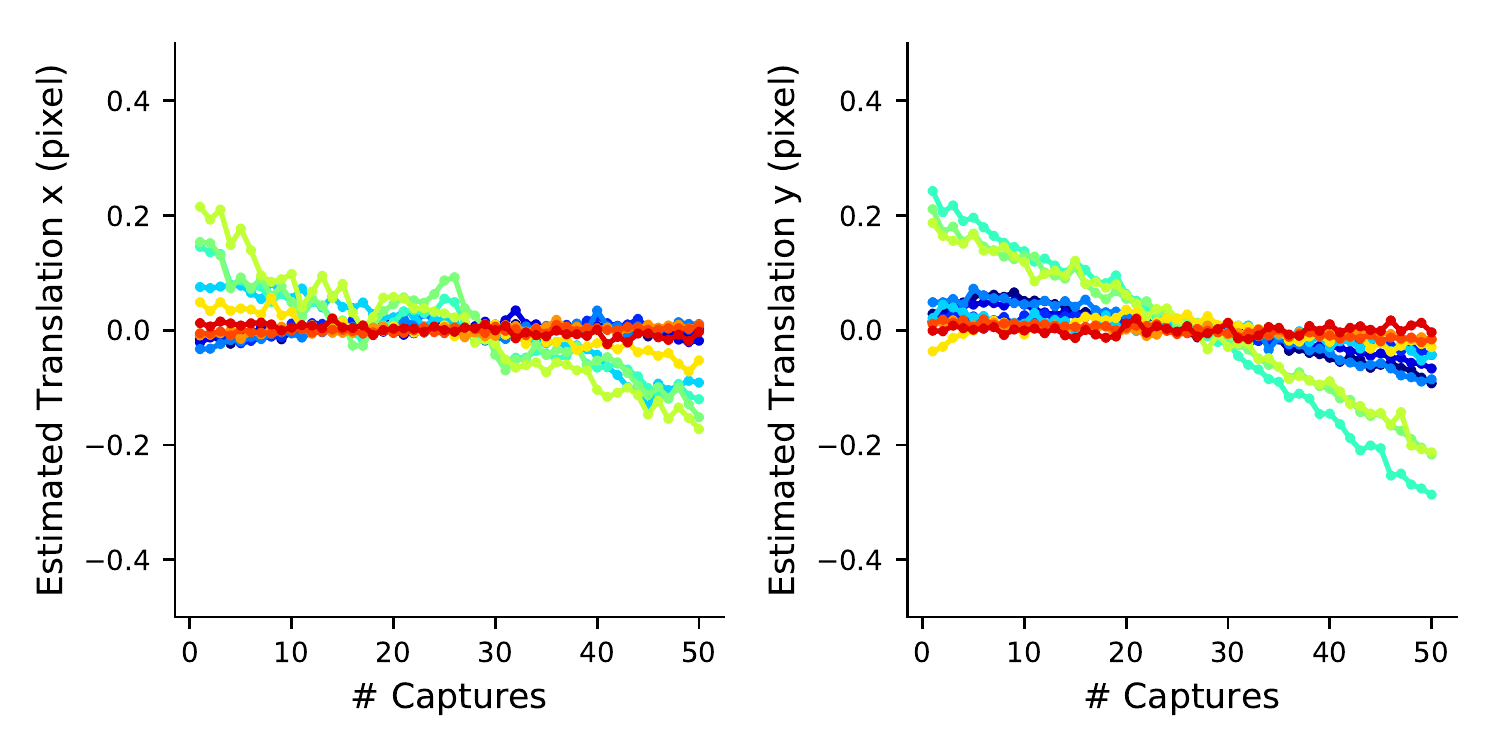}
    \caption{Estimated translation along x and y axes, both within a half-pixel (0.5). The estimation is performed on the 20-th FOV of each imaging configuration. Each line in a plot shows the estimation of one of the 12 configurations (different modalities on different samples).}
    \label{fig:est_translation}
\end{figure}

The approach to estimate ground truth by averaging a sequence of captures usually comes with the issue of spatial misalignment, which is typical in photos taken by smartphones and DSLR. We use intensity-based image registration to register a sequence of image with the same FOV against the mean image of the sequence, but find that the estimated global translations in both x and y axis are less than a half-pixel (0.5), as shown in Figure~\ref{fig:est_translation}. Translation in sub-pixel smooths out noisy images, and thus destroys the realness of Poisson noise which is the main characteristic of our dataset. In short, the image sequence obtained by the commercial fluorescence microscopes is already well aligned; thus image registration is not performed.

\paragraph{Different noise levels}
As described in Section~\ref{sec:image_acquisition}, the raw images are acquired with a low excitation power thus a relatively high noise level (low SNR) to increase the difficulty of denoising task. Meanwhile, the raw images with high noise levels allow us to create images with lower noise levels by image averaging. Particularly, we obtain averaged images with four extra noise levels by averaging $S$ ($S=2, 4, 8, 16$) raw images, respectively, within the same sequence (FOV) of 50.
We sequentially select each image within the sequence; for each selected image, $S-1$ images next to it are circularly selected; the $S$ selected images in total are used for averaging.
Using this circular averaging method, we are able to obtain the same number of averaged images as the number of raw images in the sequence, i.e., 50; meanwhile, the newly generated 50 raw images can be considered as 50 different noise realizations. In this way, the amount of noisy images in the dataset can be increased to five-fold ($S=1, 2, 4, 8, 16$).
Some example images with different noise levels are shown in Figure~\ref{fig:fig_diff_avg_roi}. 
As also shown in Table~\ref{table:benchmark_mixed_test}, the peak signal-to-noise ratio (PSNR) of the averaged images increases as the number of raw images used for averaging increases.


\paragraph{Ground truth estimation}
We estimate the ground truth by averaging all 50 captures on the same FOV, similar to the approaches employed in \cite{alexander2016registration} and \cite{luisier2011purelet}; hence in the FMD dataset, each FOV has only one ground truth that is shared by all noise realizations from that FOV.
As demonstrated in \cite{alexander2016registration} and also shown in Section~\ref{subsec:estimation_stat}, the image quality or noise characteristics of a fluorescence microscopy image will see little improvement after including around 40 images in the average; therefore, we choose 50 captures as our criterion to obtain the ground truth.
As shown in Equations~(\ref{eq:averaging_detail}) and (\ref{eq:averaging_mean_var}), the ground truth $y_i$ for images with different noise levels $z_i^j$ is the same, and image averaging is equivalent to sampling from a Poisson-Gaussian distribution with a higher SNR. Regardless of the number of images used for averaging, the mean stays the same and equals to the ground truth.
Figure~\ref{fig:fig_diff_avg_roi} shows two ground truth images as well as their corresponding noise realizations. 

\subsection{Dataset Statistics and Noise Estimation}
\label{subsec:estimation_stat}

\begin{table}[t]
\centering
\begin{tabular}{llcc}
\toprule

\multicolumn{1}{c}{Modality} &
\multicolumn{1}{c}{Samples} &
\multicolumn{1}{c}{a} &
\multicolumn{1}{c}{b} \\

\hline

\multicolumn{1}{c}{CF} &
\multicolumn{1}{c}{BPAE (Nuclei)} &
\multicolumn{1}{c}{1.39$\times10^{-2}$}&
\multicolumn{1}{c}{-2.16$\times10^{-4}$}\\

\multicolumn{1}{c}{CF} &
\multicolumn{1}{c}{BPAE (F-actin)} &
\multicolumn{1}{c}{1.37$\times10^{-2}$}&
\multicolumn{1}{c}{-1.85$\times10^{-4}$}\\

\multicolumn{1}{c}{CF} &
\multicolumn{1}{c}{BPAE (Mito)} &
\multicolumn{1}{c}{1.21$\times10^{-2}$}&
\multicolumn{1}{c}{-1.54$\times10^{-4}$}\\

\multicolumn{1}{c}{CF} &
\multicolumn{1}{c}{Zebrafish} &
\multicolumn{1}{c}{9.43$\times10^{-2}$}&
\multicolumn{1}{c}{-1.60$\times10^{-3}$}\\

\multicolumn{1}{c}{CF} &
\multicolumn{1}{c}{Mouse Brain} &
\multicolumn{1}{c}{1.94$\times10^{-2}$}&
\multicolumn{1}{c}{-2.68$\times10^{-4}$}\\

\multicolumn{1}{c}{TP} &
\multicolumn{1}{c}{BPAE (Nuclei)} &
\multicolumn{1}{c}{3.31$\times10^{-2}$}&
\multicolumn{1}{c}{-8.39$\times10^{-4}$}\\

\multicolumn{1}{c}{TP} &
\multicolumn{1}{c}{BPAE (F-actin)} &
\multicolumn{1}{c}{2.55$\times10^{-2}$}&
\multicolumn{1}{c}{-5.43$\times10^{-4}$}\\

\multicolumn{1}{c}{TP} &
\multicolumn{1}{c}{BPAE (Mito)} &
\multicolumn{1}{c}{2.10$\times10^{-2}$}&
\multicolumn{1}{c}{-4.57$\times10^{-4}$}\\

\multicolumn{1}{c}{TP} &
\multicolumn{1}{c}{Mouse Brain} &
\multicolumn{1}{c}{3.38$\times10^{-2}$}&
\multicolumn{1}{c}{-9.16$\times10^{-4}$}\\

\multicolumn{1}{c}{WF} &
\multicolumn{1}{c}{BPAE (Nuclei)} &
\multicolumn{1}{c}{2.29$\times10^{-4}$}&
\multicolumn{1}{c}{2.35$\times10^{-4}$}\\

\multicolumn{1}{c}{WF} &
\multicolumn{1}{c}{BPAE (F-actin)} &
\multicolumn{1}{c}{1.94$\times10^{-3}$}&
\multicolumn{1}{c}{1.91$\times10^{-4}$}\\

\multicolumn{1}{c}{WF} &
\multicolumn{1}{c}{BPAE (Mito)} &
\multicolumn{1}{c}{3.55$\times10^{-4}$}&
\multicolumn{1}{c}{1.95$\times10^{-4}$}\\

\bottomrule
\end{tabular}
\\
\vspace{.08cm}
\caption{Estimation of noise parameters ($a$, $b$) of the FMD dataset. The shown $a$ and $b$ are average estimation values of 20 raw noisy images from 20 different FOVs (one raw image from each FOV). CF, confocal; TP: two-photon; WF: wide-field.}
\label{tab:estimate_ab}
\vspace{-.25cm}
\end{table}

\begin{figure}[t]
\begin{center}
\includegraphics[width=1\linewidth]{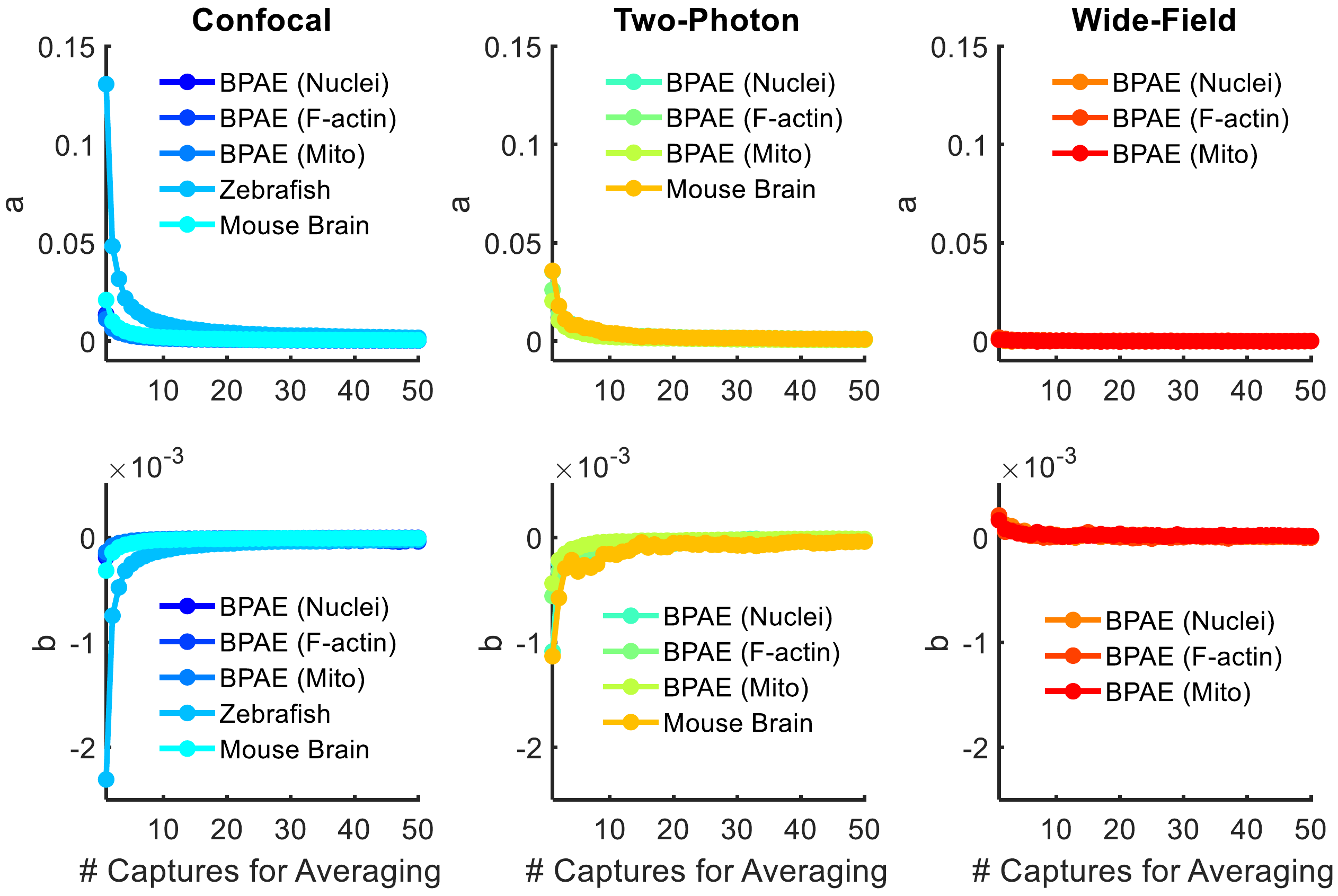}
\end{center}
\caption{Estimated noise parameters ($a$ and $b$) of averaged images obtained with different raw image numbers in the average. The estimation is performed on the second FOV of each imaging configuration.}
\label{fig:estimate_ab_stat}
\end{figure}

Taking the combination of each sample (the BPAE cells are considered as three samples due to its fluorophore composition) and each microscopy modality as a configuration, the FMD dataset includes 12 different imaging configurations that are representative of almost all fluorescence microscopy applications in practice.
For each configuration, we capture 20 different FOVs of the sample, and for each FOV, we acquire 50 raw images. Meanwhile, the 50 raw images in a FOV can be extended to five-fold using the circular averaging method described in Section~\ref{subsec:noisy_gt_estimate}. Therefore, in total, the dataset has $12 \times 20 = 240$ FOVs or ground truth images, $240 \times 50 = 12,000$ raw images, and $12,000 \times 5=60,000$ noisy images as noise realizations.

While there are blind denoising methods (e.g., DnCNN) that are able to denoise an image without any additional information, most denoising algorithms such as NLM and BM3D, however, require an estimate of the noise levels presented in the image. In this work, we employ the noise estimation method in \cite{foi2008noisemodel} to estimate the Poisson-Gaussian noise parameters, $a$ and $b$, described in Section~\ref{sec:modeling}.
The estimated values of $a$ and $b$ not only are needed in the benchmark of various denoising algorithms, they also reflect the characteristics of the noise presented in an images. Specifically, since Poisson-Gaussian noise is a mixture of both Poisson and Gaussian noises, which are parameterized by $a$ and $b$, respectively, an image with a large estimate value of $a$ but a small $b$ may be considered as a Poisson noise dominated image, while a small $a$ with a large $b$ can indicate that the image is Gaussian noise dominated. In fluorescence microscopy, however, it is unlikely to have a Gaussian noise dominated image due to the low signal levels; most fluorescence microscopy images are Poisson noise, or shot noise, dominated, with certain types of microscopes, such as wide-field ones, have a considerable amount of Gaussian noise involved \cite{bal2012denoise,meiniel2018microscopy}.
Note that the noise estimation program from \cite{foi2008noisemodel} could generate a negative $b$ value when the Gaussian noise component is small relative to the pedestal level (offset-from-zero of output). This, however, does not mean that the image has a ``negative" Gaussian noise variance. More details can be found in \cite{foi2008noisemodel}. In practice, when $b$ is estimated to be negative, we make it zero in the subsequent PURE-LET and VST-based algorithms.

We evaluate the noise characteristics of our FMD dataset by estimating the noise parameters of raw noisy image (1 in each FOV, 240 in total). The estimated $a$ and $b$ are then grouped according to their corresponding imaging configurations (20 FOVs in each configuration, 12 configurations in total) and averaged. The results are presented in Table~\ref{tab:estimate_ab}. For confocal and two-photon microscopy, the estimated $a$ are comparably large while the $b$ are negative; hence confocal and two-photon images are Poisson noise dominated. For wide-field microscopy, however, the $a$ are much smaller than above, possibly due to the much lower sensitivity of CCD cameras used in wide-field microscopy compared to the PMTs used in confocal and two-photon microscopy; meanwhile, the $b$ are now all positive, which indicates that wide-field images have a mixed Poisson-Gaussian noise with a considerable amount of Gaussian noise presented. 
We further evaluate the effect of image averaging on its noise characteristics. Figure~\ref{fig:estimate_ab_stat} shows the estimated $a$ and $b$ values when different number of images, $S$, are included in the average. The results are in good agreement with the theory in Equation~\ref{eq:averaging_mean_var} and the observations in Table~\ref{tab:estimate_ab}, as the estimated parameters follow the trend of $a/S$ and $b/S$, and their initial values ($S=1$) are close to the ones in Table~\ref{tab:estimate_ab}.
Figure~\ref{fig:estimate_ab_stat} also shows that the values of $a$ and $b$ exhibit little change when the number of captures used for averaging is more than 40; this confirms the observation reported in \cite{alexander2016registration} that the image quality or noise characteristics of a fluorescence microscopy image will see little improvement after including around 40 images in the average.

\section{Benchmark}
\label{sec:benchmark}
In this Section we benchmark several representative denoising methods, including deep learning models, on our fluorescence microscopy images with real Poisson-Gaussian noise. We show that deep learning models perform better than traditional methods on the FMD dataset.

\subsection{Setup}
The FMD dataset is split to training and test sets, where the test set is composed of images randomly selected from the 19-th FOV of each imaging configuration and noise levels (the rest 19 FOVs are for training and validation purposes). The mixed test set consists of 4 images randomly selected from the 19-th FOV of 12 imaging configurations (combination of microscopy modalities and biological samples), organized in different noise levels. Thus we have 5 mixed test sets each of which have 48 noisy images with a specific noise level corresponding to 1 (raw), 2, 4, 8, and 16 times averaging. We also test the denoising algorithms on all 50 images from the same FOV (19-th) of a specific imaging configuration, also organized in different noise levels, with denoising results shown in the supplementary material.

Considering GPU memory constraint for training fully convolutional networks~\cite{zhang2017beyond, lehtinen2018noise2noise} on large images, we crop the raw images of size $512\times 512$ to four non-overlapping patches of size $256 \times 256$. We evaluate the computation time on Intel Xeon CPU E5-2680, and additionally on Nvidia GeForce GTX 1080 Ti GPU for deep learning models.

The 10 benchmarked algorithms in this work can be divided into three categories. The first category is for the methods that are specifically designed for Poisson-Gaussian denoising; we benchmark PURE-LET \cite{luisier2011purelet}, an effective and representative Poisson-Gaussian denoising algorithm. The second category is for using well-studied Gaussian denoising methods in combination with VST and inverse VST; we combine GAT and the exact unbiased inverse transformation with classical denoising algorithms including NLM \cite{buades2005non}, BM3D \cite{dabov2007bm3d}, KSVD and its two variants KSVD(D) (over-complete DCT dictionary) and KSVD(G) (global or given dictionary) \cite{aharon2006ksvd}, EPLL \cite{zoran2011learning}, and WNNM \cite{gu2014weighted}. The last category is for deep learning based methods; we benchmark DnCNN \cite{zhang2017beyond} and Noise2Noise \cite{lehtinen2018noise2noise}.
Note that the estimation of noise parameters $a$ (scaling coefficient) and $b$ (Gaussian noise variance) are required for the algorithms in the first and second categories to work. The estimation is performed according to Section~\ref{subsec:estimation_stat} and then the images as well as the estimated parameters are sent to the denoising algorithms.

For benchmarking deep learning methods, unlike previous work~\cite{abdelhamed2018high} that directly tests with the pre-trained models, we re-train these models with the same network architecture and similar hyper-parameters on the FMD dataset from scratch.
Specifically, we compare two representative models, one of which requires ground truth (DnCNN) and the other does not (Noise2Noise).


\begin{table*}[t!]
\centering
\begin{tabular}{lcccccc}
\toprule

\multicolumn{1}{l}{} &
\multicolumn{5}{c}{Number of raw images for averaging} &\\
\cline{2-6}

\multicolumn{1}{l}{Methods} &
\multicolumn{1}{c}{1} &
\multicolumn{1}{c}{2} &
\multicolumn{1}{c}{4} &
\multicolumn{1}{c}{8} &
\multicolumn{1}{c}{16} &
\multicolumn{1}{c}{Time} \\
\hline

\multicolumn{1}{l}{Raw} &
\multicolumn{1}{c}{27.22~/~0.5442}&
\multicolumn{1}{c}{30.08~/~0.6800}&
\multicolumn{1}{c}{32.86~/~0.7981}&
\multicolumn{1}{c}{36.03~/~0.8892}&
\multicolumn{1}{c}{39.70~/~0.9487}&
\multicolumn{1}{c}{-}\\

\multicolumn{1}{l}{VST+NLM \cite{buades2005non}} &
\multicolumn{1}{c}{31.25~/~0.7503}&
\multicolumn{1}{c}{32.85~/~0.8116}&
\multicolumn{1}{c}{34.92~/~0.8763}&
\multicolumn{1}{c}{37.09~/~0.9208}&
\multicolumn{1}{c}{40.04~/~0.9540}&
\multicolumn{1}{c}{137.10 s}\\

\multicolumn{1}{l}{VST+BM3D \cite{makitalo2013vst}} &
\multicolumn{1}{c}{32.71~/~0.7922}&
\multicolumn{1}{c}{34.09~/~0.8430}&
\multicolumn{1}{c}{36.05~/~0.8970}&
\multicolumn{1}{c}{38.01~/~0.9336}&
\multicolumn{1}{c}{40.61~/~0.9598}&
\multicolumn{1}{c}{5.67 s}\\

\multicolumn{1}{l}{VST+KSVD \cite{aharon2006ksvd}} &
\multicolumn{1}{c}{32.02~/~0.7746}&
\multicolumn{1}{c}{33.69~/~0.8327}&
\multicolumn{1}{c}{35.84~/~0.8933}&
\multicolumn{1}{c}{37.79~/~0.9314}&
\multicolumn{1}{c}{40.36~/~0.9585}&
\multicolumn{1}{c}{341.21 s}\\

\multicolumn{1}{l}{VST+KSVD(D) \cite{aharon2006ksvd}} &
\multicolumn{1}{c}{31.77~/~0.7712}&
\multicolumn{1}{c}{33.45~/~0.8292}&
\multicolumn{1}{c}{35.67~/~0.8908}&
\multicolumn{1}{c}{37.69~/~0.9300}&
\multicolumn{1}{c}{40.32~/~0.9579}&
\multicolumn{1}{c}{67.96 s}\\

\multicolumn{1}{l}{VST+KSVD(G) \cite{aharon2006ksvd}} &
\multicolumn{1}{c}{31.98~/~0.7752}&
\multicolumn{1}{c}{33.64~/~0.8327}&
\multicolumn{1}{c}{35.83~/~0.8930}&
\multicolumn{1}{c}{37.82~/~0.9312}&
\multicolumn{1}{c}{40.44~/~0.9584}&
\multicolumn{1}{c}{58.82 s}\\

\multicolumn{1}{l}{VST+EPLL \cite{zoran2011learning}} &
\multicolumn{1}{c}{32.61~/~0.7876}&
\multicolumn{1}{c}{34.07~/~0.8414}&
\multicolumn{1}{c}{36.08~/~0.8970}&
\multicolumn{1}{c}{38.12~/~0.9349}&
\multicolumn{1}{c}{40.83~/~0.9618}&
\multicolumn{1}{c}{288.63 s}\\

\multicolumn{1}{l}{VST+WNNM \cite{gu2014weighted}} &
\multicolumn{1}{c}{32.52~/~0.7880}&
\multicolumn{1}{c}{34.04~/~0.8419}&
\multicolumn{1}{c}{36.04~/~0.8973}&
\multicolumn{1}{c}{37.95~/~0.9334}&
\multicolumn{1}{c}{40.45~/~0.9587}&
\multicolumn{1}{c}{451.89 s}\\

\multicolumn{1}{l}{PURE-LET \cite{luisier2011purelet}} &
\multicolumn{1}{c}{31.95~/~0.7664}&
\multicolumn{1}{c}{33.49~/~0.8270}&
\multicolumn{1}{c}{35.29~/~0.8814}&
\multicolumn{1}{c}{37.25~/~0.9212}&
\multicolumn{1}{c}{39.59~/~0.9450}&
\multicolumn{1}{c}{\textbf{2.61 s}}\\
\multicolumn{1}{l}{DnCNN \cite{zhang2017beyond}} &
\multicolumn{1}{c}{34.88~/~0.9063}&
\multicolumn{1}{c}{36.02~/~\textbf{0.9257}}&
\multicolumn{1}{c}{37.57~/~0.9460}&
\multicolumn{1}{c}{39.28~/~0.9588}&
\multicolumn{1}{c}{\textbf{41.57}~/~0.9721}&
\multicolumn{1}{c}{3.07 s$^\dagger$}\\
\multicolumn{1}{l}{Noise2Noise \cite{lehtinen2018noise2noise}} &
\multicolumn{1}{c}{\textbf{35.40}~/~\textbf{0.9187}}&
\multicolumn{1}{c}{\textbf{36.40}~/~0.9230}&
\multicolumn{1}{c}{\textbf{37.59~/~0.9481}}&
\multicolumn{1}{c}{\textbf{39.43~/~0.9601}}&
\multicolumn{1}{c}{41.45~/~\textbf{0.9724}}&
\multicolumn{1}{c}{2.94 s$^\dagger$}\\

\bottomrule
\end{tabular}
\\
\vspace{.08cm}
\caption{Denoising performance using the mixed test set, which includes confocal, two-photon, and wide-field microscopy images. PSNR (dB), SSIM, and denoising time (seconds) are obtained by averaging over 48 noise realizations in the mixed test set for each of 5 noise levels. Results of DnCNN and Noise2Noise are obtained by training on dataset with all noise levels. All 50 captures of each FOV (except the 19-th FOV which is reserved for test) are included in the training set, with 1 (DnCNN) or 2 (Noise2Noise) samples of which randomly selected from each FOV when forming mini-batches during training for 400 epochs. $^\dagger$Note that test time for deep learning models on GPU is faster in orders of magnitude, i.e. 0.62 ms for DnCNN and 0.99 ms for Noise2Noise on single GPU in our experiment.}
\label{table:benchmark_mixed_test}
\vspace{-.25cm}
\end{table*}

\subsection{Results and Discussion}

The benchmark denoising results on the mixed test set is shown in Table~\ref{table:benchmark_mixed_test}, including PSNR, structural similarity index (SSIM)~\cite{wang2004image} and denoising time.
From the table, BM3D (in combination with VST) is still the most versatile traditional denoising algorithm regarding its high PSNR and relatively fast denoising speed. PURE-LET, though its PSNR is not the highest, is the fastest denoising method among all the benchmarked algorithms thanks to its specific design for Poisson-Gaussian denoising.
Finally, deep learning models outperform the other 8 methods by a significant margin in all noise levels, both in terms of PSNR and SSIM, even thought they are \textit{blind} to noise levels. This is different from the observation made before in~\cite{abdelhamed2018high, plotz2017benchmarking}, probably because the nature of Poisson dominated noise is different from Gaussian noise while most of the denoising methods are developed for Gaussian noise model. Even if we applied the VST before Gaussian denoising, the transformed noise may still be different from a pure Gaussian one. More importantly, here the models are re-trained with our FMD dataset instead of pre-trained on other datasets.



The training data for deep learning models includes all imaging configurations and noise levels; thus we use one trained model to perform blind denoising on various imaging configurations and noise levels.
We confirm that overall the Noise2Noise model has similar denoising performance as DnCNN, but without the need of clean images, and with almost 2dB higher than VST+BM3D in PSNR \cite{lehtinen2018noise2noise}. It even performs slightly better than DnCNN in the high noise domain, which is desirable in practice. 

\begin{figure}
    \centering
    \includegraphics[width=\linewidth]{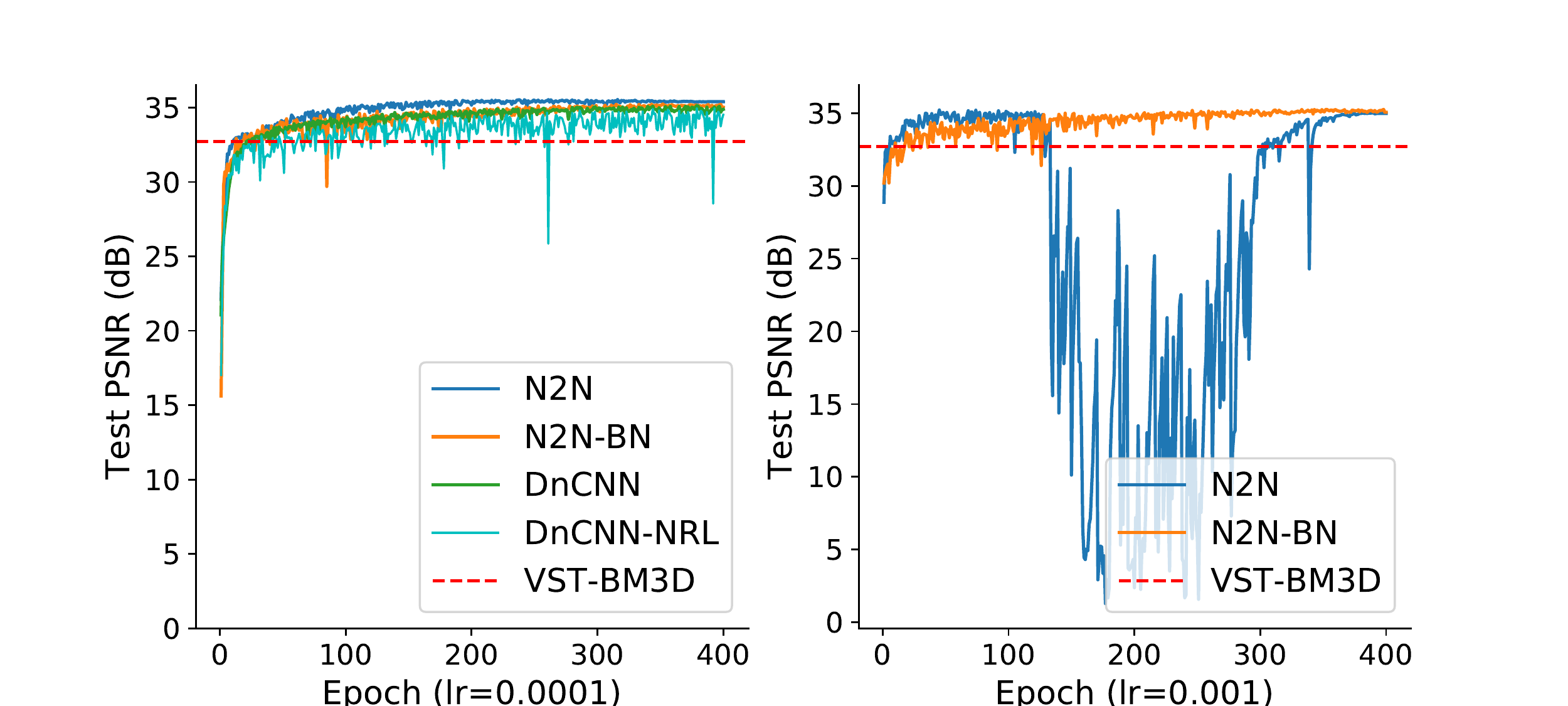}
    \caption{Test PSNR on the mixed test set with raw images during training. Each training epoch contains 18240 ($5\times 12 \times 19 \times 16$) images of size $256\times 256$.
    Given enough training time (e.g. 400 epochs), Noise2Noise eventually outperforms DnCNN and VST-BM3D. Batch normalization helps stabilize training for Noise2Noise, and for DnCNN, residual learning does help improve denoising.}
    \label{fig:cnn_training}
\end{figure}
\begin{table}[t]
\centering
\caption{PSNR (dB) on raw images in the mixed test set for the models trained with different learning rate.}
\begin{tabular}{l  c  c  c  c  c}
\toprule
Learn. Rate & 1E-3  & 5E-4  & 1E-4  & 5E-5  & 1E-5      \\
\hline
DnCNN       & 34.61 & -     & 34.88 & 34.62 & 34.01        \\
N2N         & 34.98 & 35.19 & 35.40 & 35.49 & 34.65        \\
N2N-BN      & 35.15 & 35.07 & 35.12 & 35.12 & 34.60        \\
DnCNN2      & 33.30 & -     & 34.35 & -     & 33.41        \\
\bottomrule
\end{tabular}
\label{tab:test_psnr_rebuttal}
\end{table}

We investigate the effect of adding batch normalization layers for the Noise2Noise model (i.e. N2N-BN in Figure~\ref{fig:cnn_training}), which does help stabilize the training process even when the learning rate is relatively large (e.g. 0.001), but does not improve PSNR when the learning rate is well turned (e.g. 0.0001 which is the learning rate for benchmark).
We also train DnCNN without residual learning (DnCNN-NRL) where the model directly outputs the denoised image instead of the residual between clean and noisy images, and confirm it is worse than the model with residual learning (DnCNN-RL), as has been reported in \cite{zhang2017beyond}. The test performance for the mixed test set with raw images during training is shown in Figure~\ref{fig:cnn_training} and the PSNR for each case is shown in Table~\ref{tab:test_psnr_rebuttal}.

\begin{figure}[t!]
\begin{center}
\includegraphics[width=\linewidth]{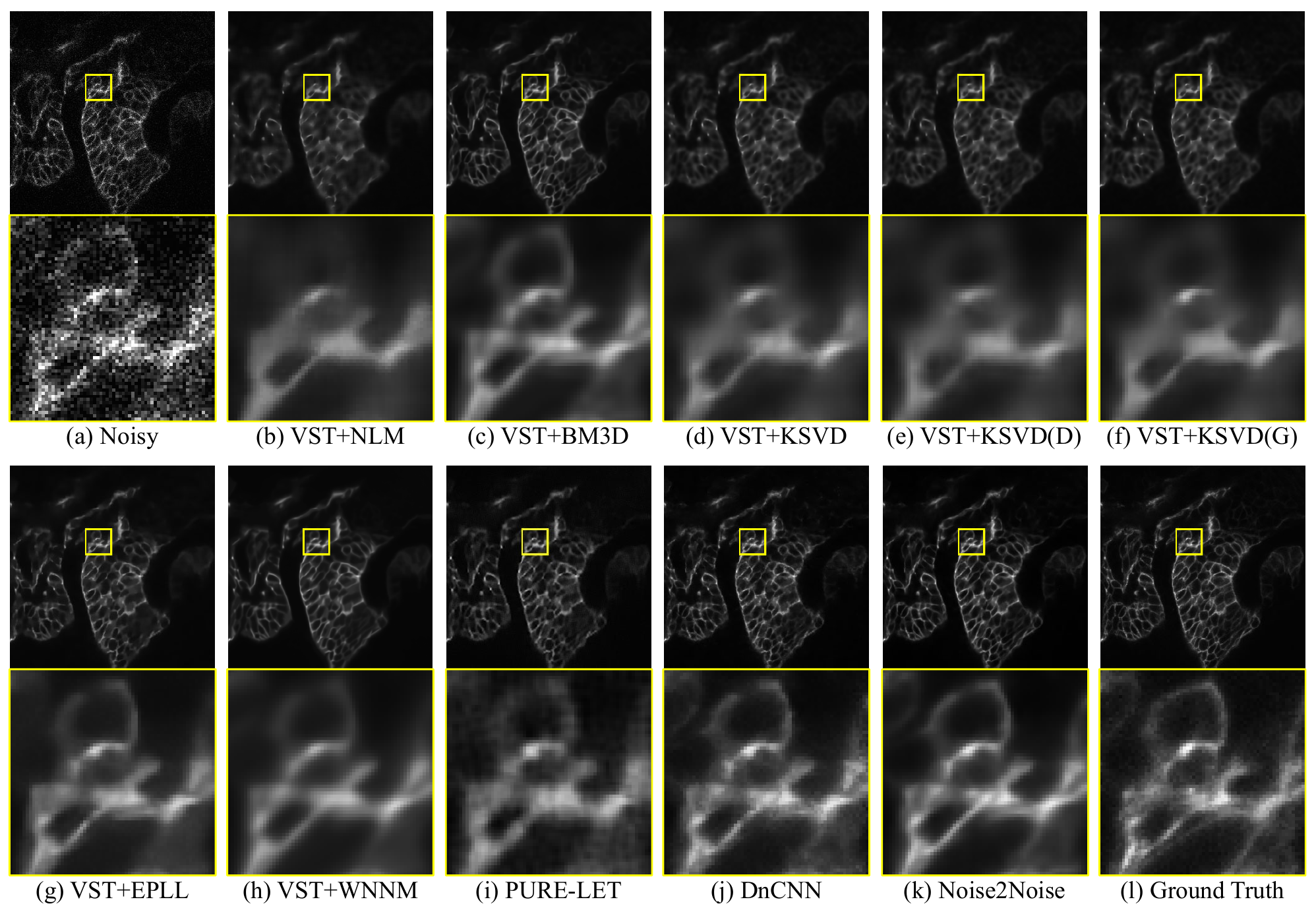}
\end{center}
\caption{Benchmark results for raw single-channel (gray) images (zebrafish embryo under confocal microscopy). PSNR and SSIM values are in Table~\ref{table:fig_caption}.}
\label{fig:benchmark_gray}
\end{figure}

\begin{figure}[t!]
\begin{center}
\includegraphics[width=1\linewidth]{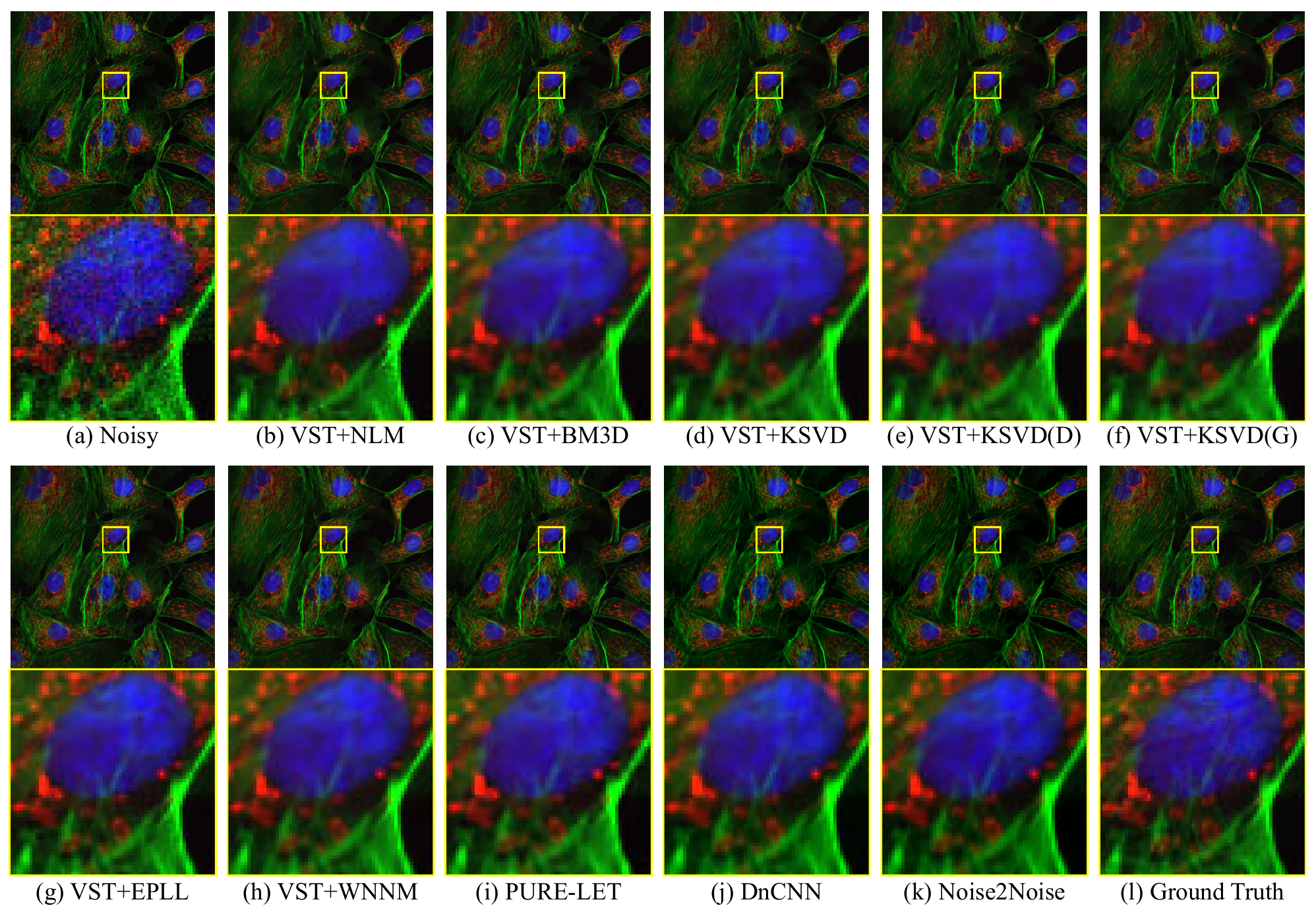}
\end{center}
\caption{Benchmark results for raw multi-channel (color) images (BPAE cells under confocal microscopy). PSNR and SSIM values are in Table~\ref{table:fig_caption}.}
\label{fig:benchmark_color}
\end{figure}

We also show benchmark results of the 10 algorithms on raw single-channel (gray) and raw multi-channel (color) confocal images in Figures~\ref{fig:benchmark_gray} and~\ref{fig:benchmark_color}, respectively, where the PSNR and SSIM of the color images are the mean values of that of their three channels.

The denoising time for deep learning models is the time to pass a mini-batch of four $256\times 256$ patches cropped from one $512\times 512$ image through the network.
Deep learning models have similar denoising time with that of VST-BM3D and PURE-LET when running on CPU. However, the denoising time can be reduced to less than 1 ms when running on GPU, which potentially enables real-time denoising up to 100 frames per second, which is out of reach of traditional denoising methods.
With such a denoising speed and high performance, deep learning denoising methods could dramatically benefit real-time fluorescence microscopy imaging, which allows biomedical researchers to observe the fast and dynamic biological processes in a much improved quality and to see processes that cannot be clearly seen before.

\begin{table}[t!]
\centering
\caption{Benchmark results [PSNR (dB) / SSIM] for confocal images of zebrafish embryo (Figure~\ref{fig:benchmark_gray}) and BPAE cells (Figure~\ref{fig:benchmark_color}).}
\begin{tabular}{lcc}
\toprule
Methods     &   Zebrafish   &   BPAE \\
\midrule
Raw         &   22.71~/~0.4441  &   30.67~/~0.7902  \\
VST+NLM     &   28.49~/~0.7952  &   34.74~/~0.9108  \\
VST+BM3D    &   31.99~/~0.8862  &   35.86~/~0.9338  \\
VST+KSVD    &   29.25~/~0.8234  &   35.72~/~0.9209  \\
VST+KSVD(D) &   29.04~/~0.8212  &   35.47~/~0.9139  \\
VST+KSVD(G) &   29.23~/~0.8232  &   35.63~/~0.9176  \\
VST+EPLL    &   31.71~/~0.8711  &   35.72~/~0.9335  \\
VST+WNNM    &   31.22~/~0.8702  &   35.89~/~0.9322  \\
PURE-LET    &   30.59~/~0.8332  &   35.18~/~0.9262  \\
DnCNN       &   32.35~/~0.8991  &   36.15~/~0.9413  \\
Noise2Noise &   \textbf{33.02~/~0.9109}  &   \textbf{36.35~/~0.9441}  \\
\bottomrule
\end{tabular}
\label{table:fig_caption}
\end{table}


\section{Conclusion}
\label{sec:conclusion}

In this work, we have constructed a dedicated denoising dataset of real fluorescence microscopy images with Poisson-Gaussian noise, which covers most microscopy modalities. We have used image averaging to obtain ground truth and noisy images with 5 different noise levels. With this dataset, we have benchmarked representative denoising algorithms for Poisson-Gaussian noise including the most recent deep learning models. The benchmark results show that deep learning denoising models trained on our FMD dataset outperforms other methods by a large margin across all imaging modalities and noise levels. We have made our FMD dataset publicly available as a benchmark for Poisson-Gaussian denoising research, which, we believe, will be especially useful for researchers that are interested in improving the imaging quality of fluorescence microscopy.

\section*{Acknowledgments}
This material is based upon work supported by the National Science Foundation under Grant No.~CBET-1554516.
Yide Zhang's research was supported by the Berry Family Foundation Graduate Fellowship of Advanced Diagnostics \& Therapeutics (AD\&T), University of Notre Dame. 
The authors further acknowledge the Notre Dame Integrated Imaging Facility (NDIIF) for the use of the Nikon A1R-MP confocal microscope and Nikon Eclipse 90i wide-field micrscope in NDIIF's Optical Microscopy Core.

{
\small
\bibliographystyle{ieee}
\bibliography{egbib}
}


\appendix
\newpage
\section{Supplementary Material}
\label{sec:sm}

\subsection{Fluorescence microscopy setup}
The confocal and two-photon images were acquired with a Nikon A1R-MP laser scanning confocal microscope equipped with a Nikon Apo LWD 40$\times$, 1.15 NA water-immersion objective. The confocal and two-photon images were 512$\times$512 pixels with a pixel size of 300 nm and a pixel dwell time of 2 $\upmu$s. The A1R-MP microscope has multiple detectors (PMTs) in parallel, so for multi-channel (color) fluorescence imaging with the BPAE cells, all three images were acquired simultaneously.
For confocal imaging, the excitation was generated by a LU4/LU4A laser unit, the pinhole size was set to 1.2 Airy unit, and the imaging conditions for different samples were as follows: BPAE nuclei, 405 nm excitation, 0.5\% laser power, 110 PMT gain; BPAE F-actin, 488 nm excitation, 0.5\% laser power, 110 PMT gain; BPAE mitochondria, 561 nm excitation, 0.5\% laser power, 110 PMT gain; mouse brain, 405 nm excitation, 0.5\% laser power, 115 PMT gain; zebrafish embryo, 488 nm excitation, 10\% laser power, 140 PMT gain.
For two-photon microscopy, the excitation was generated by a Spectra-Physics Mai Tai DeepSee femtosecond laser, and for all two-photon images, the laser power was set to 0.5\%, the PMT gain to 130, and the excitation wavelength to 780 nm. Note that our dataset did not include two-photon images of the zebrafish sample because during two-photon imaging, very strong two-photon auto-fluorescence signals from the zebrafish were observed, which severely degraded the imaging quality.

The wide-field images were acquired with a Nikon Eclipse 90i wide-field fluorescence microscope equipped with a Nikon Plan Fluor 40$\times$, 0.75 NA objective. The excitation was generated by a halogen lamp (with ND16 neutral-density filter) and the images were captured by a DS-Fi1-U2 camera with an exposure time of 200 ms and a gain of 46. The raw image size was 1280$\times$960 and the pixel size was 170 nm. These images were cropped to 512$\times$512 before being processed for our dataset.
Note that our dataset only covered wide-field images of the BPAE cells because wide-field microscopy could not image well in animal tissues such as mouse brain and zebrafish embryo, where strong out-of-focus fluorescence would blur out the wide-field images. Since the BPAE cells were stained with three different fluorophores while only one detector (CCD camera) was available in the 90i microscope, we imaged three times for the same FOV, each time with a different filter block (DAPI for nuclei, FITC for F-actin, TRITC for mitochondria), to acquire the multi-channel (color) fluorescence image of the cells.

\subsection{Pixel clipping or over/under-exposure}
In fluorescence microscopy, under-exposure is not an issue due to the high sensitivity and accuracy of microscopy detectors. 
However, pixel clipping or over-exposure could be inevitable because distinct biological structures with various optical properties could generate extremely bright fluorescence signals, which saturated the detector and caused pixel clipping.  
We tried to avoid pixel clipping by manually adjusting the detector gain. As a result, at most $0.2\%$ of pixels were clipped in all imaging configurations, as shown in Table~\ref{tab:clip_percentage} (averaged percentages).
Consequently, the clipped pixels could introduce bias when we estimated the ground truth by image averaging. Considering the negligible proportion of clipped pixels, our ground truth images maintain an accuracy higher than $99.8\%$.

\begin{table}[h]
\centering
\caption{Percentages of clipped pixels to all pixels in the images.}
\begin{tabular}{c c c c}
\hline\hline

\multicolumn{1}{c|}{Mod.} &
\multicolumn{1}{c|}{Samples} &
\multicolumn{1}{c|}{Raw (\%)} &
\multicolumn{1}{c}{GT (\%)} \\

\cline{1-4}

\multicolumn{1}{c|}{CF} &
\multicolumn{1}{c|}{BPAE (Nuclei)} &
\multicolumn{1}{c|}{0.002343}&
\multicolumn{1}{c}{0}\\

\multicolumn{1}{c|}{CF} &
\multicolumn{1}{c|}{BPAE (F-actin)} &
\multicolumn{1}{c|}{0.004214}&
\multicolumn{1}{c}{0.000629}\\

\multicolumn{1}{c|}{CF} &
\multicolumn{1}{c|}{BPAE (Mito)} &
\multicolumn{1}{c|}{0.000013}&
\multicolumn{1}{c}{0}\\

\multicolumn{1}{c|}{CF} &
\multicolumn{1}{c|}{Zebrafish} &
\multicolumn{1}{c|}{0.186157}&
\multicolumn{1}{c}{0.038757}\\

\multicolumn{1}{c|}{CF} &
\multicolumn{1}{c|}{Mouse Brain} &
\multicolumn{1}{c|}{0.015899}&
\multicolumn{1}{c}{0.000057}\\

\multicolumn{1}{c|}{TP} &
\multicolumn{1}{c|}{BPAE (Nuclei)} &
\multicolumn{1}{c|}{0.169477}&
\multicolumn{1}{c}{0.001450}\\

\multicolumn{1}{c|}{TP} &
\multicolumn{1}{c|}{BPAE (F-actin)} &
\multicolumn{1}{c|}{0.006969}&
\multicolumn{1}{c}{0.000515}\\

\multicolumn{1}{c|}{TP} &
\multicolumn{1}{c|}{BPAE (Mito)} &
\multicolumn{1}{c|}{0.000346}&
\multicolumn{1}{c}{0.000172}\\

\multicolumn{1}{c|}{TP} &
\multicolumn{1}{c|}{Mouse Brain} &
\multicolumn{1}{c|}{0.151986}&
\multicolumn{1}{c}{0.008736}\\

\multicolumn{1}{c|}{WF} &
\multicolumn{1}{c|}{BPAE (Nuclei)} &
\multicolumn{1}{c|}{0.123395}&
\multicolumn{1}{c}{0.000153}\\

\multicolumn{1}{c|}{WF} &
\multicolumn{1}{c|}{BPAE (F-actin)} &
\multicolumn{1}{c|}{0.000311}&
\multicolumn{1}{c}{0}\\

\multicolumn{1}{c|}{WF} &
\multicolumn{1}{c|}{BPAE (Mito)} &
\multicolumn{1}{c|}{0.000037}&
\multicolumn{1}{c}{0}\\

\hline\hline
\end{tabular}
\label{tab:clip_percentage}
\end{table}

\subsection{Benchmark results on separate test set}
Here we show the benchmark results on the 19-th FOV (which is pre-selected as the test set) for each imaging configuration and each noise level, which contains 50 noise realizations in each case. The results are organized in Table~\ref{table:benchmark_confocal} (confocal), \ref{table:benchmark_two_photon} (two-photon), and \ref{table:benchmark_wide_field} (wide-field). For all test cases, deep learning based denoising methods almost dominate over traditional methods.

\begin{table*}[t]
\centering
\begin{tabular}{llcccccc}

\hline\hline

\multicolumn{2}{l|}{\bf Confocal Microscopy} &
\multicolumn{5}{c||}{Number of raw images for averaging} &\\

\cline{1-7}

\multicolumn{1}{l|}{Samples} &
\multicolumn{1}{l|}{Methods} &
\multicolumn{1}{c|}{1} &
\multicolumn{1}{c|}{2} &
\multicolumn{1}{c|}{4} &
\multicolumn{1}{c|}{8} &
\multicolumn{1}{c||}{16} &
\multicolumn{1}{c}{Time} \\

\cline{1-8}

\multicolumn{1}{l|}{} &
\multicolumn{1}{l|}{VST+NLM} &
\multicolumn{1}{c|}{37.35~/~0.9656}&
\multicolumn{1}{c|}{38.20~/~0.9730}&
\multicolumn{1}{c|}{39.31~/~0.9810}&
\multicolumn{1}{c|}{41.11~/~0.9862}&
\multicolumn{1}{c||}{43.68~/~0.9906}&
\multicolumn{1}{c}{129.92 s}\\

\multicolumn{1}{l|}{} &
\multicolumn{1}{l|}{VST+BM3D} &
\multicolumn{1}{c|}{38.45~/~0.9732}&
\multicolumn{1}{c|}{39.59~/~0.9786}&
\multicolumn{1}{c|}{40.95~/~0.9853}&
\multicolumn{1}{c|}{42.37~/~0.9889}&
\multicolumn{1}{c||}{44.37~/~0.9918}&
\multicolumn{1}{c}{5.13 s}\\

\multicolumn{1}{l|}{} &
\multicolumn{1}{l|}{VST+KSVD} &
\multicolumn{1}{c|}{38.15~/~0.9699}&
\multicolumn{1}{c|}{39.48~/~0.9773}&
\multicolumn{1}{c|}{40.92~/~0.9850}&
\multicolumn{1}{c|}{42.30~/~0.9888}&
\multicolumn{1}{c||}{44.32~/~0.9919}&
\multicolumn{1}{c}{65.90 s}\\

\multicolumn{1}{l|}{} &
\multicolumn{1}{l|}{VST+KSVD(D)} &
\multicolumn{1}{c|}{37.77~/~0.9679}&
\multicolumn{1}{c|}{39.26~/~0.9762}&
\multicolumn{1}{c|}{40.82~/~0.9846}&
\multicolumn{1}{c|}{42.28~/~0.9887}&
\multicolumn{1}{c||}{44.28~/~0.9918}&
\multicolumn{1}{c}{20.25 s}\\

\multicolumn{1}{l|}{BPAE} &
\multicolumn{1}{l|}{VST+KSVD(G)} &
\multicolumn{1}{c|}{38.07~/~0.9694}&
\multicolumn{1}{c|}{39.39~/~0.9767}&
\multicolumn{1}{c|}{40.87~/~0.9847}&
\multicolumn{1}{c|}{42.28~/~0.9886}&
\multicolumn{1}{c||}{44.20~/~0.9917}&
\multicolumn{1}{c}{17.11 s}\\

\multicolumn{1}{l|}{(Nuclei)} & 
\multicolumn{1}{l|}{VST+EPLL} &
\multicolumn{1}{c|}{38.38~/~0.9731}&
\multicolumn{1}{c|}{39.47~/~0.9785}&
\multicolumn{1}{c|}{40.85~/~0.9854}&
\multicolumn{1}{c|}{42.35~/~0.9891}&
\multicolumn{1}{c||}{44.42~/~0.9920}&
\multicolumn{1}{c}{246.47 s}\\

\multicolumn{1}{l|}{} &
\multicolumn{1}{l|}{VST+WNNM} &
\multicolumn{1}{c|}{38.43~/~0.9734}&
\multicolumn{1}{c|}{39.55~/~0.9784}&
\multicolumn{1}{c|}{40.91~/~0.9851}&
\multicolumn{1}{c|}{42.28~/~0.9885}&
\multicolumn{1}{c||}{44.21~/~0.9914}&
\multicolumn{1}{c}{417.71 s}\\

\multicolumn{1}{l|}{} &
\multicolumn{1}{l|}{PURE-LET} &
\multicolumn{1}{c|}{37.15~/~0.9583}&
\multicolumn{1}{c|}{38.55~/~0.9688}&
\multicolumn{1}{c|}{40.15~/~0.9795}&
\multicolumn{1}{c|}{41.55~/~0.9843}&
\multicolumn{1}{c||}{43.51~/~0.9887}&
\multicolumn{1}{c}{2.43 s}\\

\multicolumn{1}{l|}{} &
\multicolumn{1}{l|}{DnCNN} &
\multicolumn{1}{c|}{38.91~/~\textbf{0.9795}}&
\multicolumn{1}{c|}{40.23~/~\textbf{0.9834}}&
\multicolumn{1}{c|}{\textbf{41.62}~/~\textbf{0.9872}}&
\multicolumn{1}{c|}{\textbf{43.07}~/~\textbf{0.9903}}&
\multicolumn{1}{c||}{\textbf{44.97}~/~\textbf{0.9930}}&
\multicolumn{1}{c}{\textbf{2.37} s}\\

\multicolumn{1}{l|}{} &
\multicolumn{1}{l|}{Noise2Noise} &
\multicolumn{1}{c|}{\textbf{39.13}~/~0.9771}&
\multicolumn{1}{c|}{\textbf{40.29}~/~0.9823}&
\multicolumn{1}{c|}{41.47~/~0.9858}&
\multicolumn{1}{c|}{42.73~/~0.9885}&
\multicolumn{1}{c||}{44.21~/~0.9907}&
\multicolumn{1}{c}{2.69 s}\\

\cline{1-8}

\multicolumn{1}{l|}{} &
\multicolumn{1}{l|}{VST+NLM} &
\multicolumn{1}{c|}{32.80~/~0.8419}&
\multicolumn{1}{c|}{34.28~/~0.8893}&
\multicolumn{1}{c|}{35.76~/~0.9237}&
\multicolumn{1}{c|}{37.37~/~0.9462}&
\multicolumn{1}{c||}{39.39~/~0.9624}&
\multicolumn{1}{c}{134.04 s}\\

\multicolumn{1}{l|}{} &
\multicolumn{1}{l|}{VST+BM3D} &
\multicolumn{1}{c|}{34.07~/~0.8880}&
\multicolumn{1}{c|}{35.38~/~0.9168}&
\multicolumn{1}{c|}{36.74~/~0.9395}&
\multicolumn{1}{c|}{38.15~/~0.9556}&
\multicolumn{1}{c||}{39.80~/~0.9675}&
\multicolumn{1}{c}{6.42 s}\\

\multicolumn{1}{l|}{} &
\multicolumn{1}{l|}{VST+KSVD} &
\multicolumn{1}{c|}{33.33~/~0.8565}&
\multicolumn{1}{c|}{34.81~/~0.8985}&
\multicolumn{1}{c|}{36.25~/~0.9291}&
\multicolumn{1}{c|}{37.65~/~0.9484}&
\multicolumn{1}{c||}{39.17~/~0.9614}&
\multicolumn{1}{c}{287.22 s}\\

\multicolumn{1}{l|}{} &
\multicolumn{1}{l|}{VST+KSVD(D)} &
\multicolumn{1}{c|}{32.88~/~0.8412}&
\multicolumn{1}{c|}{34.49~/~0.8892}&
\multicolumn{1}{c|}{36.07~/~0.9245}&
\multicolumn{1}{c|}{37.55~/~0.9460}&
\multicolumn{1}{c||}{39.11~/~0.9598}&
\multicolumn{1}{c}{64.16 s}\\

\multicolumn{1}{l|}{BPAE} &
\multicolumn{1}{l|}{VST+KSVD(G)} &
\multicolumn{1}{c|}{33.08~/~0.8465}&
\multicolumn{1}{c|}{34.62~/~0.8914}&
\multicolumn{1}{c|}{36.14~/~0.9248}&
\multicolumn{1}{c|}{37.60~/~0.9457}&
\multicolumn{1}{c||}{39.17~/~0.9595}&
\multicolumn{1}{c}{47.82 s}\\

\multicolumn{1}{l|}{(F-actin)} & 
\multicolumn{1}{l|}{VST+EPLL} &
\multicolumn{1}{c|}{34.07~/~0.8892}&
\multicolumn{1}{c|}{35.49~/~0.9207}&
\multicolumn{1}{c|}{36.94~/~0.9441}&
\multicolumn{1}{c|}{38.48~/~0.9604}&
\multicolumn{1}{c||}{40.35~/~0.9725}&
\multicolumn{1}{c}{317.13 s}\\

\multicolumn{1}{l|}{} &
\multicolumn{1}{l|}{VST+WNNM} &
\multicolumn{1}{c|}{33.94~/~0.8809}&
\multicolumn{1}{c|}{35.29~/~0.9126}&
\multicolumn{1}{c|}{36.59~/~0.9362}&
\multicolumn{1}{c|}{37.84~/~0.9515}&
\multicolumn{1}{c||}{39.21~/~0.9621}&
\multicolumn{1}{c}{415.91 s}\\

\multicolumn{1}{l|}{} &
\multicolumn{1}{l|}{PURE-LET} &
\multicolumn{1}{c|}{33.50~/~0.8776}&
\multicolumn{1}{c|}{34.75~/~0.9066}&
\multicolumn{1}{c|}{35.98~/~0.9283}&
\multicolumn{1}{c|}{37.16~/~0.9433}&
\multicolumn{1}{c||}{38.18~/~0.9505}&
\multicolumn{1}{c}{2.66 s}\\

\multicolumn{1}{l|}{} &
\multicolumn{1}{l|}{DnCNN} &
\multicolumn{1}{c|}{34.21~/~\textbf{0.9029}}&
\multicolumn{1}{c|}{35.62~/~\textbf{0.9311}}&
\multicolumn{1}{c|}{\textbf{37.07}~/~\textbf{0.9512}}&
\multicolumn{1}{c|}{\textbf{38.66}~/~\textbf{0.9665}}&
\multicolumn{1}{c||}{\textbf{40.75}~/~\textbf{0.9791}}&
\multicolumn{1}{c}{\textbf{2.39} s}\\

\multicolumn{1}{l|}{} &
\multicolumn{1}{l|}{Noise2Noise} &
\multicolumn{1}{c|}{\textbf{34.33}~/~0.9025}&
\multicolumn{1}{c|}{\textbf{35.63}~/~0.9289}&
\multicolumn{1}{c|}{36.92~/~0.9480}&
\multicolumn{1}{c|}{38.30~/~0.9625}&
\multicolumn{1}{c||}{39.92~/~0.9736}&
\multicolumn{1}{c}{2.58 s}\\

\cline{1-8}

\multicolumn{1}{l|}{} &
\multicolumn{1}{l|}{VST+NLM} &
\multicolumn{1}{c|}{35.79~/~0.9279}&
\multicolumn{1}{c|}{37.27~/~0.9518}&
\multicolumn{1}{c|}{38.93~/~0.9673}&
\multicolumn{1}{c|}{40.89~/~0.9781}&
\multicolumn{1}{c||}{43.36~/~0.9865}&
\multicolumn{1}{c}{130.14 s}\\

\multicolumn{1}{l|}{} &
\multicolumn{1}{l|}{VST+BM3D} &
\multicolumn{1}{c|}{37.43~/~0.9489}&
\multicolumn{1}{c|}{38.82~/~0.9632}&
\multicolumn{1}{c|}{40.27~/~0.9742}&
\multicolumn{1}{c|}{41.80~/~0.9817}&
\multicolumn{1}{c||}{43.78~/~0.9879}&
\multicolumn{1}{c}{5.92 s}\\

\multicolumn{1}{l|}{} &
\multicolumn{1}{l|}{VST+KSVD} &
\multicolumn{1}{c|}{36.97~/~0.9378}&
\multicolumn{1}{c|}{38.49~/~0.9575}&
\multicolumn{1}{c|}{39.98~/~0.9712}&
\multicolumn{1}{c|}{41.48~/~0.9798}&
\multicolumn{1}{c||}{43.33~/~0.9865}&
\multicolumn{1}{c}{241.33 s}\\

\multicolumn{1}{l|}{} &
\multicolumn{1}{l|}{VST+KSVD(D)} &
\multicolumn{1}{c|}{36.55~/~0.9305}&
\multicolumn{1}{c|}{38.25~/~0.9537}&
\multicolumn{1}{c|}{39.89~/~0.9695}&
\multicolumn{1}{c|}{41.50~/~0.9792}&
\multicolumn{1}{c||}{43.42~/~0.9864}&
\multicolumn{1}{c}{60.91 s}\\

\multicolumn{1}{l|}{BPAE} &
\multicolumn{1}{l|}{VST+KSVD(G)} &
\multicolumn{1}{c|}{36.93~/~0.9368}&
\multicolumn{1}{c|}{38.59~/~0.9579}&
\multicolumn{1}{c|}{40.18~/~0.9720}&
\multicolumn{1}{c|}{41.71~/~0.9806}&
\multicolumn{1}{c||}{43.59~/~0.9871}&
\multicolumn{1}{c}{42.51 s}\\

\multicolumn{1}{l|}{(Mito)} & 
\multicolumn{1}{l|}{VST+EPLL} &
\multicolumn{1}{c|}{37.56~/~0.9515}&
\multicolumn{1}{c|}{38.95~/~0.9653}&
\multicolumn{1}{c|}{40.41~/~0.9757}&
\multicolumn{1}{c|}{41.94~/~0.9828}&
\multicolumn{1}{c||}{43.98~/~0.9887}&
\multicolumn{1}{c}{312.86 s}\\

\multicolumn{1}{l|}{} &
\multicolumn{1}{l|}{VST+WNNM} &
\multicolumn{1}{c|}{37.46~/~0.9486}&
\multicolumn{1}{c|}{38.91~/~0.9638}&
\multicolumn{1}{c|}{40.34~/~0.9745}&
\multicolumn{1}{c|}{41.80~/~0.9816}&
\multicolumn{1}{c||}{43.67~/~0.9875}&
\multicolumn{1}{c}{502.87 s}\\

\multicolumn{1}{l|}{} &
\multicolumn{1}{l|}{PURE-LET} &
\multicolumn{1}{c|}{36.87~/~0.9433}&
\multicolumn{1}{c|}{38.12~/~0.9568}&
\multicolumn{1}{c|}{39.47~/~0.9678}&
\multicolumn{1}{c|}{40.95~/~0.9764}&
\multicolumn{1}{c||}{42.73~/~0.9834}&
\multicolumn{1}{c}{2.70 s}\\

\multicolumn{1}{l|}{} &
\multicolumn{1}{l|}{DnCNN} &
\multicolumn{1}{c|}{\textbf{37.89}~/~\textbf{0.9586}}&
\multicolumn{1}{c|}{\textbf{39.30}~/~\textbf{0.9702}}&
\multicolumn{1}{c|}{\textbf{40.68}~/~\textbf{0.9781}}&
\multicolumn{1}{c|}{\textbf{42.14}~/~\textbf{0.9841}}&
\multicolumn{1}{c||}{\textbf{44.00}~/~\textbf{0.9894}}&
\multicolumn{1}{c}{\textbf{2.38} s}\\

\multicolumn{1}{l|}{} &
\multicolumn{1}{l|}{Noise2Noise} &
\multicolumn{1}{c|}{37.74~/~0.9549}&
\multicolumn{1}{c|}{39.13~/~0.9675}&
\multicolumn{1}{c|}{40.47~/~0.9756}&
\multicolumn{1}{c|}{41.78~/~0.9813}&
\multicolumn{1}{c||}{43.22~/~0.9859}&
\multicolumn{1}{c}{2.59 s}\\

\cline{1-8}

\multicolumn{1}{l|}{} &
\multicolumn{1}{l|}{VST+NLM} &
\multicolumn{1}{c|}{28.23~/~0.7895}&
\multicolumn{1}{c|}{31.47~/~0.8593}&
\multicolumn{1}{c|}{34.00~/~0.9078}&
\multicolumn{1}{c|}{35.72~/~0.9328}&
\multicolumn{1}{c||}{37.58~/~0.9482}&
\multicolumn{1}{c}{145.64 s}\\

\multicolumn{1}{l|}{} &
\multicolumn{1}{l|}{VST+BM3D} &
\multicolumn{1}{c|}{32.00~/~0.8854}&
\multicolumn{1}{c|}{33.75~/~0.9102}&
\multicolumn{1}{c|}{35.30~/~0.9301}&
\multicolumn{1}{c|}{36.78~/~0.9443}&
\multicolumn{1}{c||}{38.32~/~0.9546}&
\multicolumn{1}{c}{6.29 s}\\

\multicolumn{1}{l|}{} &
\multicolumn{1}{l|}{VST+KSVD} &
\multicolumn{1}{c|}{29.04~/~0.8203}&
\multicolumn{1}{c|}{32.17~/~0.8740}&
\multicolumn{1}{c|}{34.58~/~0.9167}&
\multicolumn{1}{c|}{36.31~/~0.9388}&
\multicolumn{1}{c||}{37.86~/~0.9519}&
\multicolumn{1}{c}{60.01 s}\\

\multicolumn{1}{l|}{} &
\multicolumn{1}{l|}{VST+KSVD(D)} &
\multicolumn{1}{c|}{28.87~/~0.8184}&
\multicolumn{1}{c|}{31.42~/~0.8647}&
\multicolumn{1}{c|}{33.97~/~0.9093}&
\multicolumn{1}{c|}{35.97~/~0.9350}&
\multicolumn{1}{c||}{37.74~/~0.9504}&
\multicolumn{1}{c}{12.54 s}\\

\multicolumn{1}{l|}{Zebrafish} &
\multicolumn{1}{l|}{VST+KSVD(G)} &
\multicolumn{1}{c|}{29.03~/~0.8201}&
\multicolumn{1}{c|}{31.88~/~0.8701}&
\multicolumn{1}{c|}{34.34~/~0.9133}&
\multicolumn{1}{c|}{36.26~/~0.9374}&
\multicolumn{1}{c||}{38.04~/~0.9520}&
\multicolumn{1}{c}{9.93 s}\\

\multicolumn{1}{l|}{Embryo} & 
\multicolumn{1}{l|}{VST+EPLL} &
\multicolumn{1}{c|}{31.62~/~0.8678}&
\multicolumn{1}{c|}{33.66~/~0.9048}&
\multicolumn{1}{c|}{35.34~/~0.9298}&
\multicolumn{1}{c|}{36.92~/~0.9460}&
\multicolumn{1}{c||}{38.61~/~0.9574}&
\multicolumn{1}{c}{317.67 s}\\

\multicolumn{1}{l|}{} &
\multicolumn{1}{l|}{VST+WNNM} &
\multicolumn{1}{c|}{30.94~/~0.8654}&
\multicolumn{1}{c|}{33.43~/~0.9048}&
\multicolumn{1}{c|}{35.23~/~0.9284}&
\multicolumn{1}{c|}{36.74~/~0.9432}&
\multicolumn{1}{c||}{38.14~/~0.9527}&
\multicolumn{1}{c}{615.40 s}\\

\multicolumn{1}{l|}{} &
\multicolumn{1}{l|}{PURE-LET} &
\multicolumn{1}{c|}{30.03~/~0.8019}&
\multicolumn{1}{c|}{32.48~/~0.8817}&
\multicolumn{1}{c|}{33.84~/~0.8960}&
\multicolumn{1}{c|}{35.65~/~0.9254}&
\multicolumn{1}{c||}{37.15~/~0.9394}&
\multicolumn{1}{c}{2.59 s}\\

\multicolumn{1}{l|}{} &
\multicolumn{1}{l|}{DnCNN} &
\multicolumn{1}{c|}{32.44~/~0.9025}&
\multicolumn{1}{c|}{34.16~/~0.9267}&
\multicolumn{1}{c|}{\textbf{35.75}~/~\textbf{0.9425}}&
\multicolumn{1}{c|}{\textbf{37.28}~/~\textbf{0.9548}}&
\multicolumn{1}{c||}{\textbf{39.07}~/~\textbf{0.9659}}&
\multicolumn{1}{c}{\textbf{2.44} s}\\

\multicolumn{1}{l|}{} &
\multicolumn{1}{l|}{Noise2Noise} &
\multicolumn{1}{c|}{\textbf{32.93}~/~\textbf{0.9076}}&
\multicolumn{1}{c|}{\textbf{34.37}~/~\textbf{0.9270}}&
\multicolumn{1}{c|}{35.71~/~0.9410}&
\multicolumn{1}{c|}{37.06~/~0.9523}&
\multicolumn{1}{c||}{38.65~/~0.9625}&
\multicolumn{1}{c}{2.68 s}\\

\cline{1-8}

\multicolumn{1}{l|}{} &
\multicolumn{1}{l|}{VST+NLM} &
\multicolumn{1}{c|}{36.31~/~0.9534}&
\multicolumn{1}{c|}{37.53~/~0.9632}&
\multicolumn{1}{c|}{38.95~/~0.9706}&
\multicolumn{1}{c|}{40.87~/~0.9763}&
\multicolumn{1}{c||}{43.37~/~0.9819}&
\multicolumn{1}{c}{131.08 s}\\

\multicolumn{1}{l|}{} &
\multicolumn{1}{l|}{VST+BM3D} &
\multicolumn{1}{c|}{37.95~/~0.9637}&
\multicolumn{1}{c|}{39.47~/~0.9704}&
\multicolumn{1}{c|}{41.09~/~0.9765}&
\multicolumn{1}{c|}{42.73~/~0.9811}&
\multicolumn{1}{c||}{44.52~/~0.9847}&
\multicolumn{1}{c}{6.24 s}\\

\multicolumn{1}{l|}{} &
\multicolumn{1}{l|}{VST+KSVD} &
\multicolumn{1}{c|}{37.46~/~0.9587}&
\multicolumn{1}{c|}{39.24~/~0.9684}&
\multicolumn{1}{c|}{40.94~/~0.9757}&
\multicolumn{1}{c|}{42.55~/~0.9807}&
\multicolumn{1}{c||}{44.24~/~0.9846}&
\multicolumn{1}{c}{85.33 s}\\

\multicolumn{1}{l|}{} &
\multicolumn{1}{l|}{VST+KSVD(D)} &
\multicolumn{1}{c|}{36.67~/~0.9544}&
\multicolumn{1}{c|}{38.68~/~0.9659}&
\multicolumn{1}{c|}{40.63~/~0.9746}&
\multicolumn{1}{c|}{42.43~/~0.9804}&
\multicolumn{1}{c||}{44.26~/~0.9846}&
\multicolumn{1}{c}{21.95 s}\\

\multicolumn{1}{l|}{Mouse} &
\multicolumn{1}{l|}{VST+KSVD(G)} &
\multicolumn{1}{c|}{37.30~/~0.9582}&
\multicolumn{1}{c|}{39.15~/~0.9681}&
\multicolumn{1}{c|}{40.93~/~0.9757}&
\multicolumn{1}{c|}{42.65~/~0.9808}&
\multicolumn{1}{c||}{44.49~/~0.9849}&
\multicolumn{1}{c}{17.89 s}\\

\multicolumn{1}{l|}{Brain} & 
\multicolumn{1}{l|}{VST+EPLL} &
\multicolumn{1}{c|}{37.92~/~0.9640}&
\multicolumn{1}{c|}{39.50~/~0.9710}&
\multicolumn{1}{c|}{41.18~/~0.9772}&
\multicolumn{1}{c|}{42.87~/~0.9818}&
\multicolumn{1}{c||}{44.73~/~0.9855}&
\multicolumn{1}{c}{320.98 s}\\

\multicolumn{1}{l|}{} &
\multicolumn{1}{l|}{VST+WNNM} &
\multicolumn{1}{c|}{37.86~/~0.9624}&
\multicolumn{1}{c|}{39.47~/~0.9698}&
\multicolumn{1}{c|}{41.08~/~0.9761}&
\multicolumn{1}{c|}{42.62~/~0.9804}&
\multicolumn{1}{c||}{44.17~/~0.9837}&
\multicolumn{1}{c}{456.09 s}\\

\multicolumn{1}{l|}{} &
\multicolumn{1}{l|}{PURE-LET} &
\multicolumn{1}{c|}{36.60~/~0.9359}&
\multicolumn{1}{c|}{38.10~/~0.9477}&
\multicolumn{1}{c|}{40.06~/~0.9650}&
\multicolumn{1}{c|}{41.75~/~0.9739}&
\multicolumn{1}{c||}{43.29~/~0.9791}&
\multicolumn{1}{c}{2.54 s}\\

\multicolumn{1}{l|}{} &
\multicolumn{1}{l|}{DnCNN} &
\multicolumn{1}{c|}{38.15~/~\textbf{0.9672}}&
\multicolumn{1}{c|}{\textbf{39.78}~/~\textbf{0.9741}}&
\multicolumn{1}{c|}{\textbf{41.41}~/~\textbf{0.9794}}&
\multicolumn{1}{c|}{\textbf{43.11}~/~\textbf{0.9841}}&
\multicolumn{1}{c||}{\textbf{45.20}~/~\textbf{0.9887}}&
\multicolumn{1}{c}{\textbf{2.35} s}\\

\multicolumn{1}{l|}{} &
\multicolumn{1}{l|}{Noise2Noise} &
\multicolumn{1}{c|}{\textbf{38.19}~/~0.9665}&
\multicolumn{1}{c|}{39.77~/~0.9735}&
\multicolumn{1}{c|}{41.28~/~0.9787}&
\multicolumn{1}{c|}{42.83~/~0.9831}&
\multicolumn{1}{c||}{44.56~/~0.9869}&
\multicolumn{1}{c}{2.71 s}\\

\hline\hline
\end{tabular}
\\
\vspace{.08cm}
\caption{Denoising performance of confocal microscopy images (the 19-th FOV of each imaging configuration). PSNR (dB), SSIM, and denoising time (seconds) are obtained by averaging over 50 noise realizations through imaging experiments.}
\vspace{-.25cm}
\label{table:benchmark_confocal}
\end{table*}

\begin{table*}[t]
\centering
\begin{tabular}{llcccccc}
\hline\hline

\multicolumn{2}{l|}{\bf Two-Photon Microscopy} &
\multicolumn{5}{c||}{Number of raw images for averaging} &\\

\cline{1-7}

\multicolumn{1}{l|}{Samples} &
\multicolumn{1}{l|}{Methods} &
\multicolumn{1}{c|}{1} &
\multicolumn{1}{c|}{2} &
\multicolumn{1}{c|}{4} &
\multicolumn{1}{c|}{8} &
\multicolumn{1}{c||}{16} &
\multicolumn{1}{c}{Time} \\

\cline{1-8}

\multicolumn{1}{l|}{} &
\multicolumn{1}{l|}{VST+NLM} &
\multicolumn{1}{c|}{31.34~/~0.9173}&
\multicolumn{1}{c|}{32.13~/~0.9286}&
\multicolumn{1}{c|}{32.95~/~0.9390}&
\multicolumn{1}{c|}{34.14~/~0.9482}&
\multicolumn{1}{c||}{37.35~/~0.9571}&
\multicolumn{1}{c}{137.27 s}\\

\multicolumn{1}{l|}{} &
\multicolumn{1}{l|}{VST+BM3D} &
\multicolumn{1}{c|}{32.02~/~0.9297}&
\multicolumn{1}{c|}{32.70~/~0.9382}&
\multicolumn{1}{c|}{33.43~/~0.9458}&
\multicolumn{1}{c|}{34.60~/~0.9526}&
\multicolumn{1}{c||}{37.77~/~0.9592}&
\multicolumn{1}{c}{5.58 s}\\

\multicolumn{1}{l|}{} &
\multicolumn{1}{l|}{VST+KSVD} &
\multicolumn{1}{c|}{31.71~/~0.9227}&
\multicolumn{1}{c|}{32.55~/~0.9352}&
\multicolumn{1}{c|}{33.37~/~0.9453}&
\multicolumn{1}{c|}{34.55~/~0.9535}&
\multicolumn{1}{c||}{37.70~/~0.9613}&
\multicolumn{1}{c}{42.51 s}\\

\multicolumn{1}{l|}{} &
\multicolumn{1}{l|}{VST+KSVD(D)} &
\multicolumn{1}{c|}{31.48~/~0.9195}&
\multicolumn{1}{c|}{32.33~/~0.9323}&
\multicolumn{1}{c|}{33.23~/~0.9438}&
\multicolumn{1}{c|}{34.48~/~0.9529}&
\multicolumn{1}{c||}{37.69~/~0.9612}&
\multicolumn{1}{c}{10.77 s}\\

\multicolumn{1}{l|}{BPAE} &
\multicolumn{1}{l|}{VST+KSVD(G)} &
\multicolumn{1}{c|}{31.70~/~0.9225}&
\multicolumn{1}{c|}{32.52~/~0.9347}&
\multicolumn{1}{c|}{33.34~/~0.9448}&
\multicolumn{1}{c|}{34.55~/~0.9533}&
\multicolumn{1}{c||}{37.75~/~0.9613}&
\multicolumn{1}{c}{8.12 s}\\

\multicolumn{1}{l|}{(Nuclei)} & 
\multicolumn{1}{l|}{VST+EPLL} &
\multicolumn{1}{c|}{32.00~/~0.9313}&
\multicolumn{1}{c|}{32.70~/~0.9404}&
\multicolumn{1}{c|}{33.48~/~0.9483}&
\multicolumn{1}{c|}{34.69~/~0.9552}&
\multicolumn{1}{c||}{37.95~/~0.9618}&
\multicolumn{1}{c}{284.32 s}\\

\multicolumn{1}{l|}{} &
\multicolumn{1}{l|}{VST+WNNM} &
\multicolumn{1}{c|}{32.01~/~0.9298}&
\multicolumn{1}{c|}{32.68~/~0.9383}&
\multicolumn{1}{c|}{33.41~/~0.9460}&
\multicolumn{1}{c|}{34.55~/~0.9524}&
\multicolumn{1}{c||}{37.62~/~0.9585}&
\multicolumn{1}{c}{487.02 s}\\

\multicolumn{1}{l|}{} &
\multicolumn{1}{l|}{PURE-LET} &
\multicolumn{1}{c|}{31.62~/~0.9101}&
\multicolumn{1}{c|}{32.27~/~0.9198}&
\multicolumn{1}{c|}{32.88~/~0.9231}&
\multicolumn{1}{c|}{33.97~/~0.9312}&
\multicolumn{1}{c||}{36.92~/~0.9439}&
\multicolumn{1}{c}{2.68 s}\\

\multicolumn{1}{l|}{} &
\multicolumn{1}{l|}{DnCNN} &
\multicolumn{1}{c|}{31.59~/~0.9250}&
\multicolumn{1}{c|}{32.46~/~0.9421}&
\multicolumn{1}{c|}{33.38~/~\textbf{0.9513}}&
\multicolumn{1}{c|}{34.75~/~\textbf{0.9598}}&
\multicolumn{1}{c||}{\textbf{38.30}~/~\textbf{0.9705}}&
\multicolumn{1}{c}{\textbf{2.16} s}\\

\multicolumn{1}{l|}{} &
\multicolumn{1}{l|}{Noise2Noise} &
\multicolumn{1}{c|}{\textbf{32.44}~/~\textbf{0.9354}}&
\multicolumn{1}{c|}{\textbf{33.21}~/~\textbf{0.9434}}&
\multicolumn{1}{c|}{\textbf{34.04}~/~0.9509}&
\multicolumn{1}{c|}{\textbf{35.19}~/~0.9590}&
\multicolumn{1}{c||}{38.22~/~0.9685}&
\multicolumn{1}{c}{2.51 s}\\

\cline{1-8}

\multicolumn{1}{l|}{} &
\multicolumn{1}{l|}{VST+NLM} &
\multicolumn{1}{c|}{30.26~/~0.7176}&
\multicolumn{1}{c|}{31.43~/~0.7799}&
\multicolumn{1}{c|}{32.70~/~0.8404}&
\multicolumn{1}{c|}{34.24~/~0.8912}&
\multicolumn{1}{c||}{37.04~/~0.9297}&
\multicolumn{1}{c}{229.93 s}\\

\multicolumn{1}{l|}{} &
\multicolumn{1}{l|}{VST+BM3D} &
\multicolumn{1}{c|}{31.59~/~0.8037}&
\multicolumn{1}{c|}{32.52~/~0.8442}&
\multicolumn{1}{c|}{33.56~/~0.8813}&
\multicolumn{1}{c|}{34.91~/~0.9139}&
\multicolumn{1}{c||}{37.56~/~0.9408}&
\multicolumn{1}{c}{5.89 s}\\

\multicolumn{1}{l|}{} &
\multicolumn{1}{l|}{VST+KSVD} &
\multicolumn{1}{c|}{30.67~/~0.7381}&
\multicolumn{1}{c|}{31.84~/~0.7992}&
\multicolumn{1}{c|}{33.10~/~0.8560}&
\multicolumn{1}{c|}{34.54~/~0.8995}&
\multicolumn{1}{c||}{37.07~/~0.9304}&
\multicolumn{1}{c}{163.48 s}\\

\multicolumn{1}{l|}{} &
\multicolumn{1}{l|}{VST+KSVD(D)} &
\multicolumn{1}{c|}{30.43~/~0.7261}&
\multicolumn{1}{c|}{31.52~/~0.7833}&
\multicolumn{1}{c|}{32.83~/~0.8438}&
\multicolumn{1}{c|}{34.38~/~0.8936}&
\multicolumn{1}{c||}{37.00~/~0.9279}&
\multicolumn{1}{c}{30.14 s}\\

\multicolumn{1}{l|}{BPAE} &
\multicolumn{1}{l|}{VST+KSVD(G)} &
\multicolumn{1}{c|}{30.57~/~0.7325}&
\multicolumn{1}{c|}{31.69~/~0.7904}&
\multicolumn{1}{c|}{32.97~/~0.8485}&
\multicolumn{1}{c|}{34.48~/~0.8952}&
\multicolumn{1}{c||}{37.09~/~0.9284}&
\multicolumn{1}{c}{24.08 s}\\

\multicolumn{1}{l|}{(F-actin)} & 
\multicolumn{1}{l|}{VST+EPLL} &
\multicolumn{1}{c|}{31.48~/~0.7950}&
\multicolumn{1}{c|}{32.56~/~0.8456}&
\multicolumn{1}{c|}{33.72~/~0.8889}&
\multicolumn{1}{c|}{35.19~/~0.9237}&
\multicolumn{1}{c||}{38.09~/~0.9507}&
\multicolumn{1}{c}{287.27 s}\\

\multicolumn{1}{l|}{} &
\multicolumn{1}{l|}{VST+WNNM} &
\multicolumn{1}{c|}{31.24~/~0.7778}&
\multicolumn{1}{c|}{32.30~/~0.8278}&
\multicolumn{1}{c|}{33.41~/~0.8723}&
\multicolumn{1}{c|}{34.76~/~0.9082}&
\multicolumn{1}{c||}{37.25~/~0.9345}&
\multicolumn{1}{c}{506.98 s}\\

\multicolumn{1}{l|}{} &
\multicolumn{1}{l|}{PURE-LET} &
\multicolumn{1}{c|}{31.19~/~0.7858}&
\multicolumn{1}{c|}{32.09~/~0.8267}&
\multicolumn{1}{c|}{33.19~/~0.8705}&
\multicolumn{1}{c|}{34.53~/~0.9055}&
\multicolumn{1}{c||}{36.85~/~0.9295}&
\multicolumn{1}{c}{2.62 s}\\

\multicolumn{1}{l|}{} &
\multicolumn{1}{l|}{DnCNN} &
\multicolumn{1}{c|}{31.52~/~0.8222}&
\multicolumn{1}{c|}{32.67~/~0.8685}&
\multicolumn{1}{c|}{33.92~/~\textbf{0.9059}}&
\multicolumn{1}{c|}{35.47~/~\textbf{0.9368}}&
\multicolumn{1}{c||}{\textbf{38.68}~/~\textbf{0.9643}}&
\multicolumn{1}{c}{\textbf{2.10} s}\\

\multicolumn{1}{l|}{} &
\multicolumn{1}{l|}{Noise2Noise} &
\multicolumn{1}{c|}{\textbf{32.00}~/~\textbf{0.8257}}&
\multicolumn{1}{c|}{\textbf{33.10}~/~\textbf{0.8701}}&
\multicolumn{1}{c|}{\textbf{34.19}~/~0.9048}&
\multicolumn{1}{c|}{\textbf{35.59}~/~0.9342}&
\multicolumn{1}{c||}{38.46~/~0.9596}&
\multicolumn{1}{c}{2.32 s}\\

\cline{1-8}

\multicolumn{1}{l|}{} &
\multicolumn{1}{l|}{VST+NLM} &
\multicolumn{1}{c|}{35.11~/~0.8525}&
\multicolumn{1}{c|}{36.73~/~0.8917}&
\multicolumn{1}{c|}{38.66~/~0.9290}&
\multicolumn{1}{c|}{40.68~/~0.9554}&
\multicolumn{1}{c||}{43.49~/~0.9738}&
\multicolumn{1}{c}{208.28 s}\\

\multicolumn{1}{l|}{} &
\multicolumn{1}{l|}{VST+BM3D} &
\multicolumn{1}{c|}{37.52~/~0.9130}&
\multicolumn{1}{c|}{38.72~/~0.9338}&
\multicolumn{1}{c|}{40.09~/~0.9511}&
\multicolumn{1}{c|}{41.62~/~0.9648}&
\multicolumn{1}{c||}{43.97~/~0.9766}&
\multicolumn{1}{c}{5.49 s}\\

\multicolumn{1}{l|}{} &
\multicolumn{1}{l|}{VST+KSVD} &
\multicolumn{1}{c|}{35.75~/~0.8679}&
\multicolumn{1}{c|}{37.34~/~0.9039}&
\multicolumn{1}{c|}{39.21~/~0.9367}&
\multicolumn{1}{c|}{40.98~/~0.9576}&
\multicolumn{1}{c||}{43.29~/~0.9725}&
\multicolumn{1}{c}{97.25 s}\\

\multicolumn{1}{l|}{} &
\multicolumn{1}{l|}{VST+KSVD(D)} &
\multicolumn{1}{c|}{35.61~/~0.8648}&
\multicolumn{1}{c|}{36.96~/~0.8961}&
\multicolumn{1}{c|}{38.77~/~0.9295}&
\multicolumn{1}{c|}{40.66~/~0.9536}&
\multicolumn{1}{c||}{43.12~/~0.9710}&
\multicolumn{1}{c}{19.32 s}\\

\multicolumn{1}{l|}{BPAE} &
\multicolumn{1}{l|}{VST+KSVD(G)} &
\multicolumn{1}{c|}{35.74~/~0.8675}&
\multicolumn{1}{c|}{37.25~/~0.9019}&
\multicolumn{1}{c|}{39.16~/~0.9354}&
\multicolumn{1}{c|}{41.07~/~0.9579}&
\multicolumn{1}{c||}{43.57~/~0.9737}&
\multicolumn{1}{c}{14.39 s}\\

\multicolumn{1}{l|}{(Mito)} & 
\multicolumn{1}{l|}{VST+EPLL} &
\multicolumn{1}{c|}{37.29~/~0.9065}&
\multicolumn{1}{c|}{38.81~/~0.9348}&
\multicolumn{1}{c|}{40.38~/~0.9549}&
\multicolumn{1}{c|}{42.05~/~0.9689}&
\multicolumn{1}{c||}{44.58~/~0.9800}&
\multicolumn{1}{c}{291.54 s}\\

\multicolumn{1}{l|}{} &
\multicolumn{1}{l|}{VST+WNNM} &
\multicolumn{1}{c|}{36.68~/~0.8929}&
\multicolumn{1}{c|}{38.30~/~0.9250}&
\multicolumn{1}{c|}{39.90~/~0.9481}&
\multicolumn{1}{c|}{41.51~/~0.9636}&
\multicolumn{1}{c||}{43.77~/~0.9754}&
\multicolumn{1}{c}{525.45 s}\\

\multicolumn{1}{l|}{} &
\multicolumn{1}{l|}{PURE-LET} &
\multicolumn{1}{c|}{36.88~/~0.8946}&
\multicolumn{1}{c|}{38.01~/~0.9179}&
\multicolumn{1}{c|}{38.70~/~0.9276}&
\multicolumn{1}{c|}{40.12~/~0.9459}&
\multicolumn{1}{c||}{42.27~/~0.9637}&
\multicolumn{1}{c}{2.77 s}\\

\multicolumn{1}{l|}{} &
\multicolumn{1}{l|}{DnCNN} &
\multicolumn{1}{c|}{\textbf{38.15}~/~\textbf{0.9251}}&
\multicolumn{1}{c|}{\textbf{39.46}~/~\textbf{0.9460}}&
\multicolumn{1}{c|}{\textbf{40.87}~/~\textbf{0.9616}}&
\multicolumn{1}{c|}{\textbf{42.51}~/~\textbf{0.9738}}&
\multicolumn{1}{c||}{\textbf{45.32}~/~\textbf{0.9845}}&
\multicolumn{1}{c}{\textbf{2.10} s}\\

\multicolumn{1}{l|}{} &
\multicolumn{1}{l|}{Noise2Noise} &
\multicolumn{1}{c|}{38.11~/~0.9241}&
\multicolumn{1}{c|}{39.38~/~0.9450}&
\multicolumn{1}{c|}{40.77~/~0.9606}&
\multicolumn{1}{c|}{42.37~/~0.9727}&
\multicolumn{1}{c||}{44.82~/~0.9825}&
\multicolumn{1}{c}{2.33 s}\\

\cline{1-8}

\multicolumn{1}{l|}{} &
\multicolumn{1}{l|}{VST+NLM} &
\multicolumn{1}{c|}{32.80~/~0.9134}&
\multicolumn{1}{c|}{33.88~/~0.9237}&
\multicolumn{1}{c|}{34.88~/~0.9317}&
\multicolumn{1}{c|}{36.31~/~0.9384}&
\multicolumn{1}{c||}{38.96~/~0.9449}&
\multicolumn{1}{c}{211.65 s}\\

\multicolumn{1}{l|}{} &
\multicolumn{1}{l|}{VST+BM3D} &
\multicolumn{1}{c|}{33.81~/~0.9246}&
\multicolumn{1}{c|}{34.78~/~0.9317}&
\multicolumn{1}{c|}{35.77~/~0.9379}&
\multicolumn{1}{c|}{36.97~/~0.9431}&
\multicolumn{1}{c||}{39.39~/~0.9481}&
\multicolumn{1}{c}{6.14 s}\\

\multicolumn{1}{l|}{} &
\multicolumn{1}{l|}{VST+KSVD} &
\multicolumn{1}{c|}{33.35~/~0.9183}&
\multicolumn{1}{c|}{34.47~/~0.9288}&
\multicolumn{1}{c|}{35.60~/~0.9374}&
\multicolumn{1}{c|}{36.85~/~0.9442}&
\multicolumn{1}{c||}{39.27~/~0.9509}&
\multicolumn{1}{c}{79.00 s}\\

\multicolumn{1}{l|}{} &
\multicolumn{1}{l|}{VST+KSVD(D)} &
\multicolumn{1}{c|}{32.89~/~0.9147}&
\multicolumn{1}{c|}{34.14~/~0.9264}&
\multicolumn{1}{c|}{35.43~/~0.9362}&
\multicolumn{1}{c|}{36.79~/~0.9437}&
\multicolumn{1}{c||}{39.26~/~0.9507}&
\multicolumn{1}{c}{13.64 s}\\

\multicolumn{1}{l|}{Mouse} &
\multicolumn{1}{l|}{VST+KSVD(G)} &
\multicolumn{1}{c|}{33.34~/~0.9179}&
\multicolumn{1}{c|}{34.50~/~0.9285}&
\multicolumn{1}{c|}{35.66~/~0.9372}&
\multicolumn{1}{c|}{36.94~/~0.9441}&
\multicolumn{1}{c||}{39.42~/~0.9508}&
\multicolumn{1}{c}{9.83 s}\\

\multicolumn{1}{l|}{Brain} & 
\multicolumn{1}{l|}{VST+EPLL} &
\multicolumn{1}{c|}{33.86~/~\textbf{0.9262}}&
\multicolumn{1}{c|}{34.86~/~\textbf{0.9339}}&
\multicolumn{1}{c|}{35.86~/~0.9403}&
\multicolumn{1}{c|}{37.11~/~0.9456}&
\multicolumn{1}{c||}{39.61~/~0.9506}&
\multicolumn{1}{c}{286.50 s}\\

\multicolumn{1}{l|}{} &
\multicolumn{1}{l|}{VST+WNNM} &
\multicolumn{1}{c|}{33.79~/~0.9254}&
\multicolumn{1}{c|}{34.75~/~0.9323}&
\multicolumn{1}{c|}{35.74~/~0.9386}&
\multicolumn{1}{c|}{36.91~/~0.9435}&
\multicolumn{1}{c||}{39.22~/~0.9480}&
\multicolumn{1}{c}{512.61 s}\\

\multicolumn{1}{l|}{} &
\multicolumn{1}{l|}{PURE-LET} &
\multicolumn{1}{c|}{32.86~/~0.8812}&
\multicolumn{1}{c|}{33.47~/~0.8720}&
\multicolumn{1}{c|}{34.42~/~0.8769}&
\multicolumn{1}{c|}{35.49~/~0.8878}&
\multicolumn{1}{c||}{37.40~/~0.8997}&
\multicolumn{1}{c}{2.84 s}\\

\multicolumn{1}{l|}{} &
\multicolumn{1}{l|}{DnCNN} &
\multicolumn{1}{c|}{33.67~/~0.9068}&
\multicolumn{1}{c|}{34.95~/~0.9290}&
\multicolumn{1}{c|}{36.10~/~\textbf{0.9413}}&
\multicolumn{1}{c|}{37.43~/~\textbf{0.9507}}&
\multicolumn{1}{c||}{\textbf{40.30}~/~\textbf{0.9630}}&
\multicolumn{1}{c}{\textbf{2.30} s}\\

\multicolumn{1}{l|}{} &
\multicolumn{1}{l|}{Noise2Noise} &
\multicolumn{1}{c|}{\textbf{34.33}~/~0.9249}&
\multicolumn{1}{c|}{\textbf{35.32}~/~0.9335}&
\multicolumn{1}{c|}{\textbf{36.25}~/~0.9410}&
\multicolumn{1}{c|}{\textbf{37.46}~/~0.9499}&
\multicolumn{1}{c||}{39.89~/~0.9609}&
\multicolumn{1}{c}{2.63 s}\\

\hline\hline
\end{tabular}
\\
\vspace{.08cm}
\caption{Denoising performance of two-photon microscopy images (the 19-th FOV of each imaging configuration). PSNR (dB), SSIM, and denoising time (seconds) are obtained by averaging over 50 noise realizations through imaging experiments.}
\vspace{-.25cm}
\label{table:benchmark_two_photon}
\end{table*}

\begin{table*}[t]
\centering
\begin{tabular}{llcccccc}
\hline\hline

\multicolumn{2}{l|}{\bf Wide-Field Microscopy} &
\multicolumn{5}{c||}{Number of raw images for averaging} &\\

\cline{1-7}

\multicolumn{1}{l|}{Samples} &
\multicolumn{1}{l|}{Methods} &
\multicolumn{1}{c|}{1} &
\multicolumn{1}{c|}{2} &
\multicolumn{1}{c|}{4} &
\multicolumn{1}{c|}{8} &
\multicolumn{1}{c||}{16} &
\multicolumn{1}{c}{Time} \\

\cline{1-8}

\multicolumn{1}{l|}{} &
\multicolumn{1}{l|}{VST+NLM} &
\multicolumn{1}{c|}{25.53~/~0.3875}&
\multicolumn{1}{c|}{28.49~/~0.5548}&
\multicolumn{1}{c|}{31.36~/~0.7122}&
\multicolumn{1}{c|}{34.33~/~0.8397}&
\multicolumn{1}{c||}{37.74~/~0.9264}&
\multicolumn{1}{c}{138.54 s}\\

\multicolumn{1}{l|}{} &
\multicolumn{1}{l|}{VST+BM3D} &
\multicolumn{1}{c|}{26.22~/~0.4339}&
\multicolumn{1}{c|}{29.16~/~0.6020}&
\multicolumn{1}{c|}{31.99~/~0.7511}&
\multicolumn{1}{c|}{34.91~/~0.8650}&
\multicolumn{1}{c||}{38.25~/~0.9386}&
\multicolumn{1}{c}{6.13 s}\\

\multicolumn{1}{l|}{} &
\multicolumn{1}{l|}{VST+KSVD} &
\multicolumn{1}{c|}{26.38~/~0.4459}&
\multicolumn{1}{c|}{29.31~/~0.6132}&
\multicolumn{1}{c|}{32.10~/~0.7577}&
\multicolumn{1}{c|}{34.99~/~0.8681}&
\multicolumn{1}{c||}{38.30~/~0.9397}&
\multicolumn{1}{c}{1348.61 s}\\

\multicolumn{1}{l|}{} &
\multicolumn{1}{l|}{VST+KSVD(D)} &
\multicolumn{1}{c|}{26.41~/~0.4489}&
\multicolumn{1}{c|}{29.33~/~0.6152}&
\multicolumn{1}{c|}{32.11~/~0.7590}&
\multicolumn{1}{c|}{35.00~/~0.8688}&
\multicolumn{1}{c||}{38.30~/~0.9398}&
\multicolumn{1}{c}{183.82 s}\\

\multicolumn{1}{l|}{BPAE} &
\multicolumn{1}{l|}{VST+KSVD(G)} &
\multicolumn{1}{c|}{26.40~/~0.4533}&
\multicolumn{1}{c|}{29.32~/~0.6182}&
\multicolumn{1}{c|}{32.10~/~0.7604}&
\multicolumn{1}{c|}{34.98~/~0.8693}&
\multicolumn{1}{c||}{38.28~/~0.9399}&
\multicolumn{1}{c}{170.94 s}\\

\multicolumn{1}{l|}{(Nuclei)} & 
\multicolumn{1}{l|}{VST+EPLL} &
\multicolumn{1}{c|}{26.06~/~0.4244}&
\multicolumn{1}{c|}{29.00~/~0.5923}&
\multicolumn{1}{c|}{31.86~/~0.7440}&
\multicolumn{1}{c|}{34.79~/~0.8601}&
\multicolumn{1}{c||}{38.15~/~0.9365}&
\multicolumn{1}{c}{354.13 s}\\

\multicolumn{1}{l|}{} &
\multicolumn{1}{l|}{VST+WNNM} &
\multicolumn{1}{c|}{26.36~/~0.4440}&
\multicolumn{1}{c|}{29.29~/~0.6116}&
\multicolumn{1}{c|}{32.11~/~0.7581}&
\multicolumn{1}{c|}{35.01~/~0.8690}&
\multicolumn{1}{c||}{38.32~/~0.9402}&
\multicolumn{1}{c}{420.74 s}\\

\multicolumn{1}{l|}{} &
\multicolumn{1}{l|}{PURE-LET} &
\multicolumn{1}{c|}{26.13~/~0.4258}&
\multicolumn{1}{c|}{29.05~/~0.5931}&
\multicolumn{1}{c|}{31.89~/~0.7442}&
\multicolumn{1}{c|}{34.79~/~0.8593}&
\multicolumn{1}{c||}{38.07~/~0.9341}&
\multicolumn{1}{c}{2.49 s}\\

\multicolumn{1}{l|}{} &
\multicolumn{1}{l|}{DnCNN} &
\multicolumn{1}{c|}{33.43~/~0.8898}&
\multicolumn{1}{c|}{35.56~/~0.9262}&
\multicolumn{1}{c|}{37.05~/~0.9437}&
\multicolumn{1}{c|}{38.40~/~0.9548}&
\multicolumn{1}{c||}{40.12~/~0.9651}&
\multicolumn{1}{c}{\textbf{2.48} s}\\

\multicolumn{1}{l|}{} &
\multicolumn{1}{l|}{Noise2Noise} &
\multicolumn{1}{c|}{\textbf{36.26}~/~\textbf{0.9409}}&
\multicolumn{1}{c|}{\textbf{37.12}~/~\textbf{0.9462}}&
\multicolumn{1}{c|}{\textbf{37.88}~/~\textbf{0.9508}}&
\multicolumn{1}{c|}{\textbf{38.80}~/~\textbf{0.9569}}&
\multicolumn{1}{c||}{\textbf{40.33}~/~\textbf{0.9660}}&
\multicolumn{1}{c}{2.64 s}\\

\cline{1-8}

\multicolumn{1}{l|}{} &
\multicolumn{1}{l|}{VST+NLM} &
\multicolumn{1}{c|}{23.93~/~0.3370}&
\multicolumn{1}{c|}{27.02~/~0.4988}&
\multicolumn{1}{c|}{30.21~/~0.6672}&
\multicolumn{1}{c|}{33.58~/~0.8096}&
\multicolumn{1}{c||}{37.67~/~0.9150}&
\multicolumn{1}{c}{132.00 s}\\

\multicolumn{1}{l|}{} &
\multicolumn{1}{l|}{VST+BM3D} &
\multicolumn{1}{c|}{24.72~/~0.3792}&
\multicolumn{1}{c|}{27.84~/~0.5467}&
\multicolumn{1}{c|}{31.02~/~0.7084}&
\multicolumn{1}{c|}{34.36~/~0.8367}&
\multicolumn{1}{c||}{38.27~/~0.9258}&
\multicolumn{1}{c}{5.66 s}\\

\multicolumn{1}{l|}{} &
\multicolumn{1}{l|}{VST+KSVD} &
\multicolumn{1}{c|}{24.94~/~0.3910}&
\multicolumn{1}{c|}{28.03~/~0.5575}&
\multicolumn{1}{c|}{31.22~/~0.7178}&
\multicolumn{1}{c|}{34.54~/~0.8426}&
\multicolumn{1}{c||}{38.48~/~-0.9292}&
\multicolumn{1}{c}{1343.88 s}\\

\multicolumn{1}{l|}{} &
\multicolumn{1}{l|}{VST+KSVD(D)} &
\multicolumn{1}{c|}{25.01~/~0.3965}&
\multicolumn{1}{c|}{28.11~/~0.5629}&
\multicolumn{1}{c|}{31.28~/~0.7213}&
\multicolumn{1}{c|}{34.59~/~0.8445}&
\multicolumn{1}{c||}{38.51~/~0.9297}&
\multicolumn{1}{c}{175.55 s}\\

\multicolumn{1}{l|}{BPAE} &
\multicolumn{1}{l|}{VST+KSVD(G)} &
\multicolumn{1}{c|}{25.04~/~0.4036}&
\multicolumn{1}{c|}{28.13~/~0.5683}&
\multicolumn{1}{c|}{31.30~/~0.7245}&
\multicolumn{1}{c|}{34.60~/~0.8458}&
\multicolumn{1}{c||}{38.50~/~0.9299}&
\multicolumn{1}{c}{156.79 s}\\

\multicolumn{1}{l|}{(F-actin)} & 
\multicolumn{1}{l|}{VST+EPLL} &
\multicolumn{1}{c|}{24.55~/~0.3711}&
\multicolumn{1}{c|}{27.70~/~0.5393}&
\multicolumn{1}{c|}{30.88~/~0.7018}&
\multicolumn{1}{c|}{34.24~/~0.8331}&
\multicolumn{1}{c||}{38.16~/~0.9241}&
\multicolumn{1}{c}{352.19 s}\\

\multicolumn{1}{l|}{} &
\multicolumn{1}{l|}{VST+WNNM} &
\multicolumn{1}{c|}{24.94~/~0.3900}&
\multicolumn{1}{c|}{28.01~/~0.5560}&
\multicolumn{1}{c|}{31.17~/~0.7154}&
\multicolumn{1}{c|}{34.48~/~0.8406}&
\multicolumn{1}{c||}{38.36~/~0.9272}&
\multicolumn{1}{c}{438.09 s}\\

\multicolumn{1}{l|}{} &
\multicolumn{1}{l|}{PURE-LET} &
\multicolumn{1}{c|}{24.67~/~0.3736}&
\multicolumn{1}{c|}{27.75~/~0.5393}&
\multicolumn{1}{c|}{30.90~/~0.7012}&
\multicolumn{1}{c|}{34.18~/~0.8306}&
\multicolumn{1}{c||}{37.64~/~0.9134}&
\multicolumn{1}{c}{2.49 s}\\

\multicolumn{1}{l|}{} &
\multicolumn{1}{l|}{DnCNN} &
\multicolumn{1}{c|}{32.54~/~0.8050}&
\multicolumn{1}{c|}{34.27~/~0.8486}&
\multicolumn{1}{c|}{35.78~/~0.8817}&
\multicolumn{1}{c|}{37.47~/~0.9133}&
\multicolumn{1}{c||}{39.62~/~0.9436}&
\multicolumn{1}{c}{\textbf{2.06} s}\\

\multicolumn{1}{l|}{} &
\multicolumn{1}{l|}{Noise2Noise} &
\multicolumn{1}{c|}{\textbf{33.30}~/~\textbf{0.8264}}&
\multicolumn{1}{c|}{\textbf{34.67}~/~\textbf{0.8590}}&
\multicolumn{1}{c|}{\textbf{36.03}~/~\textbf{0.8869}}&
\multicolumn{1}{c|}{\textbf{37.65}~/~\textbf{0.9162}}&
\multicolumn{1}{c||}{\textbf{39.75}~/~\textbf{0.9452}}&
\multicolumn{1}{c}{2.66 s}\\

\cline{1-8}

\multicolumn{1}{l|}{} &
\multicolumn{1}{l|}{VST+NLM} &
\multicolumn{1}{c|}{26.26~/~0.4134}&
\multicolumn{1}{c|}{29.35~/~0.5850}&
\multicolumn{1}{c|}{32.55~/~0.7418}&
\multicolumn{1}{c|}{35.96~/~0.8610}&
\multicolumn{1}{c||}{39.93~/~0.9389}&
\multicolumn{1}{c}{134.42 s}\\

\multicolumn{1}{l|}{} &
\multicolumn{1}{l|}{VST+BM3D} &
\multicolumn{1}{c|}{26.93~/~0.4611}&
\multicolumn{1}{c|}{30.03~/~0.6312}&
\multicolumn{1}{c|}{33.24~/~0.7778}&
\multicolumn{1}{c|}{36.65~/~0.8831}&
\multicolumn{1}{c||}{40.58~/~0.9487}&
\multicolumn{1}{c}{5.97 s}\\  

\multicolumn{1}{l|}{} &
\multicolumn{1}{l|}{VST+KSVD} &
\multicolumn{1}{c|}{27.11~/~0.4737}&
\multicolumn{1}{c|}{30.20~/~0.6417}&
\multicolumn{1}{c|}{33.38~/~0.7845}&
\multicolumn{1}{c|}{36.76~/~0.8863}&
\multicolumn{1}{c||}{40.70~/~0.9504}&
\multicolumn{1}{c}{1247.01 s}\\

\multicolumn{1}{l|}{} &
\multicolumn{1}{l|}{VST+KSVD(D)} &
\multicolumn{1}{c|}{27.14~/~0.4768}&
\multicolumn{1}{c|}{30.22~/~0.6440}&
\multicolumn{1}{c|}{33.40~/~0.7859}&
\multicolumn{1}{c|}{36.78~/~0.8869}&
\multicolumn{1}{c||}{40.69~/~0.9504}&
\multicolumn{1}{c}{172.92 s}\\

\multicolumn{1}{l|}{BPAE} &
\multicolumn{1}{l|}{VST+KSVD(G)} &
\multicolumn{1}{c|}{27.13~/~0.4804}&
\multicolumn{1}{c|}{30.22~/~0.6464}&
\multicolumn{1}{c|}{33.40~/~0.7870}&
\multicolumn{1}{c|}{36.76~/~0.8872}&
\multicolumn{1}{c||}{40.66~/~0.9503}&
\multicolumn{1}{c}{161.30 s}\\

\multicolumn{1}{l|}{(Mito)} & 
\multicolumn{1}{l|}{VST+EPLL} &
\multicolumn{1}{c|}{26.80~/~0.4524}&
\multicolumn{1}{c|}{29.91~/~0.6233}&
\multicolumn{1}{c|}{33.12~/~0.7721}&
\multicolumn{1}{c|}{36.51~/~0.8791}&
\multicolumn{1}{c||}{40.46~/~0.9471}&
\multicolumn{1}{c}{345.91 s}\\

\multicolumn{1}{l|}{} &
\multicolumn{1}{l|}{VST+WNNM} &
\multicolumn{1}{c|}{27.08~/~0.4709}&
\multicolumn{1}{c|}{30.17~/~0.6400}&
\multicolumn{1}{c|}{33.37~/~0.7841}&
\multicolumn{1}{c|}{36.77~/~0.8866}&
\multicolumn{1}{c||}{40.69~/~0.9502}&
\multicolumn{1}{c}{430.37 s}\\

\multicolumn{1}{l|}{} &
\multicolumn{1}{l|}{PURE-LET} &
\multicolumn{1}{c|}{26.85~/~0.4528}&
\multicolumn{1}{c|}{29.94~/~0.6231}&
\multicolumn{1}{c|}{33.13~/~0.7709}&
\multicolumn{1}{c|}{36.49~/~0.8777}&
\multicolumn{1}{c||}{40.27~/~0.9440}&
\multicolumn{1}{c}{2.56 s}\\

\multicolumn{1}{l|}{} &
\multicolumn{1}{l|}{DnCNN} &
\multicolumn{1}{c|}{34.87~/~0.8965}&
\multicolumn{1}{c|}{36.90~/~0.9228}&
\multicolumn{1}{c|}{38.75~/~0.9405}&
\multicolumn{1}{c|}{40.65~/~0.9552}&
\multicolumn{1}{c||}{42.78~/~0.9684}&
\multicolumn{1}{c}{\textbf{2.18} s}\\

\multicolumn{1}{l|}{} &
\multicolumn{1}{l|}{Noise2Noise} &
\multicolumn{1}{c|}{\textbf{35.55}~/~\textbf{0.9105}}&
\multicolumn{1}{c|}{\textbf{37.30~/~0.9288}}&
\multicolumn{1}{c|}{\textbf{39.08~/~0.9436}}&
\multicolumn{1}{c|}{\textbf{40.88~/~0.9567}}&
\multicolumn{1}{c||}{\textbf{42.91~/~0.9692}}&
\multicolumn{1}{c}{2.71 s}\\

\hline\hline
\end{tabular}
\\
\vspace{.08cm}
\caption{Denoising performance of wide-field microscopy images (the 19-th FOV of each imaging configuration). PSNR (dB), SSIM, and denoising time (seconds) are obtained by averaging over 50 noise realizations through imaging experiments.}
\vspace{-.25cm}
\label{table:benchmark_wide_field}
\end{table*}

\subsection{Network architecture and training details}
\textbf{Network} We try our best to maintain the same network structure of DnCNN and Noise2Noise as the original papers. For N2N-BN model, we modify the Noise2Noise model by inserting batch normalization layer after each convolution layer and adding Tanh activation before the network output. For more details, please refer to the official implementations of DnCNN\footnote{\url{https://github.com/cszn/DnCNN}} and Noise2Noise\footnote{\url{https://github.com/NVlabs/noise2noise}}.

\textbf{Training} Input images are of size $256 \times 256$, normalized to the range $[-0.5, 0.5]$. Adam optimizer is used with hyperparameters $\beta_1=0.9, \beta_2=0.99$, weight decay $0.0$. 
The learning rate scheduling follows the one cycle policy\footnote{\url{https://github.com/fastai/fastai/blob/master/fastai/callbacks/one_cycle.py}}, with maximum learning rate to be $0.0001$, initial learning rate to be $1/10$ of the maximum rate, then linearly increasing the learning rate to the maximum within $0.3$ of the total epochs, then cosine annealing of the learning rate to $1/10^5$ of the maximum learning rate. The model is trained for 400 epochs. All the settings above are the same for both DnCNN and Noise2Noise.

The minibatch size is $16$ for both DnCNN and Noise2Noise. We randomly sample 4 noisy images for DnCNN (4 pairs of large noisy images for Noise2Noise) of size $512\times 512$ from the training set and crop each large image into 4 non-overlapping patches of size $256 \times 256$, thus the mini-batch size is actually 16.

\end{document}